\documentclass[10pt,journal,compsoc,twoside]{IEEEtran}
\ifCLASSOPTIONcompsoc
  \usepackage[nocompress]{cite}
\else
  \usepackage{cite}
\fi

\ifCLASSINFOpdf
  \usepackage[pdftex]{graphicx}
  \graphicspath{{./Pics/}}
  \DeclareGraphicsExtensions{.pdf,.jpeg,.png}
\else
\fi
\usepackage{indentfirst}
\usepackage{amsmath}
\usepackage[switch,columnwise]{lineno}

\usepackage{algorithmic}
\usepackage{array}
\ifCLASSOPTIONcompsoc
  \usepackage[caption=false,font=footnotesize,labelfont=sf,textfont=sf]{subfig}
\else
  \usepackage[caption=false,font=footnotesize]{subfig}
\fi
\usepackage{fixltx2e}

\usepackage[hidelinks]{hyperref}

\usepackage{epstopdf}

\usepackage{tikz}
\newcommand*\circled[1]{\tikz[baseline=(char.base)]{
  \node[shape=circle,draw,inner sep=1pt] (char) {#1};}}

\usepackage{multirow}
\usepackage{ragged2e}
\usepackage{tabularx}
\usepackage{amsfonts}

\begin{document}
%
\title{Piecewise Flat Embedding for Image Segmentation}
%
%
%
%

\author{Chaowei Fang,~\IEEEmembership{}
        Zicheng Liao,~\IEEEmembership{}
        Yizhou Yu~\IEEEmembership{}
\IEEEcompsocitemizethanks{\IEEEcompsocthanksitem C. Fang and Y. Yu are with the Department of Computer Science, The University of Hong Kong, Hong Kong. Z. Liao is with the College of Computer Science, Zhejiang University in Hangzhou, China. \protect\\
E-mail: chwfang@connect.hku.hk; zliao@zju.edu.cn; yizhouy@acm.org
}
}

\IEEEtitleabstractindextext{%
\begin{abstract}
We introduce a new multi-dimensional nonlinear embedding -- Piecewise Flat Embedding (PFE) -- for image segmentation.
Based on the theory of sparse signal recovery, piecewise flat embedding with diverse channels attempts to recover a piecewise constant image representation with sparse region boundaries and sparse cluster value scattering. The resultant piecewise flat embedding exhibits interesting properties such as suppressing slowly varying signals, and offers an image representation with higher region identifiability which is desirable for image segmentation or high-level semantic analysis tasks.
We formulate our embedding as a variant of the Laplacian Eigenmap embedding with an $L_{1,p} (0<p\leq1)$  regularization term to promote sparse solutions. First, we devise a two-stage numerical algorithm based on Bregman iterations to compute $L_{1,1}$-regularized piecewise flat embeddings. We further generalize this algorithm through iterative reweighting to solve the general $L_{1,p}$-regularized problem. To demonstrate its efficacy, we integrate PFE into two existing image segmentation frameworks, segmentation based on clustering and hierarchical segmentation based on contour detection. Experiments on four major benchmark datasets, \emph{BSDS500}, \emph{MSRC}, \emph{Stanford Background Dataset}, and \emph{PASCAL Context}, show that segmentation algorithms incorporating our embedding achieve significantly improved results.
\end{abstract}

}

\maketitle

\IEEEdisplaynontitleabstractindextext

%
\IEEEpeerreviewmaketitle

\IEEEraisesectionheading{\section{Introduction}\label{sec:introduction}}
\IEEEPARstart{I}MAGE segmentation decomposes an image into non-overlapping regions each of which covers pixels sharing common low-level(e.g. texture and color) or high-level(e.g. semantic meaning) characteristics. 
It supports high-level visual inference and perception~\cite{marr1982computational} such as figure/ground analysis and object discovery.
It supplies vital cues to methods that generate object proposals~\cite{pont2017multiscale}, which often serve as the starting point for object detection. It is also important for low-level operations, such as object extraction~\cite{fan2001automatic}, image compositing~\cite{kim2016salient} and photo enhancement~\cite{BiHY15}.

Many challenges exist in image segmentation, including the existence of textures and low-frequency appearance variations over objects in an image. Textures bring in high frequency variations (i.e. the Retinex theory~\cite{Land:71}); while the low-frequency appearance variations are caused by multiple factors including spatially varying illumination and shading. Both textures and low-frequency variations could make the differences between distant pixels on the same object exceed the differences between nearby pixels on different objects, increasing the difficulty to decide where object boundaries should be. While there have been solutions for suppressing textures using local cues~\cite{martin2004learning}, low-frequency variations are harder to remove and would require more global solutions.

\begin{figure}[t]
    \centering
    \includegraphics[width=\columnwidth,page=1]{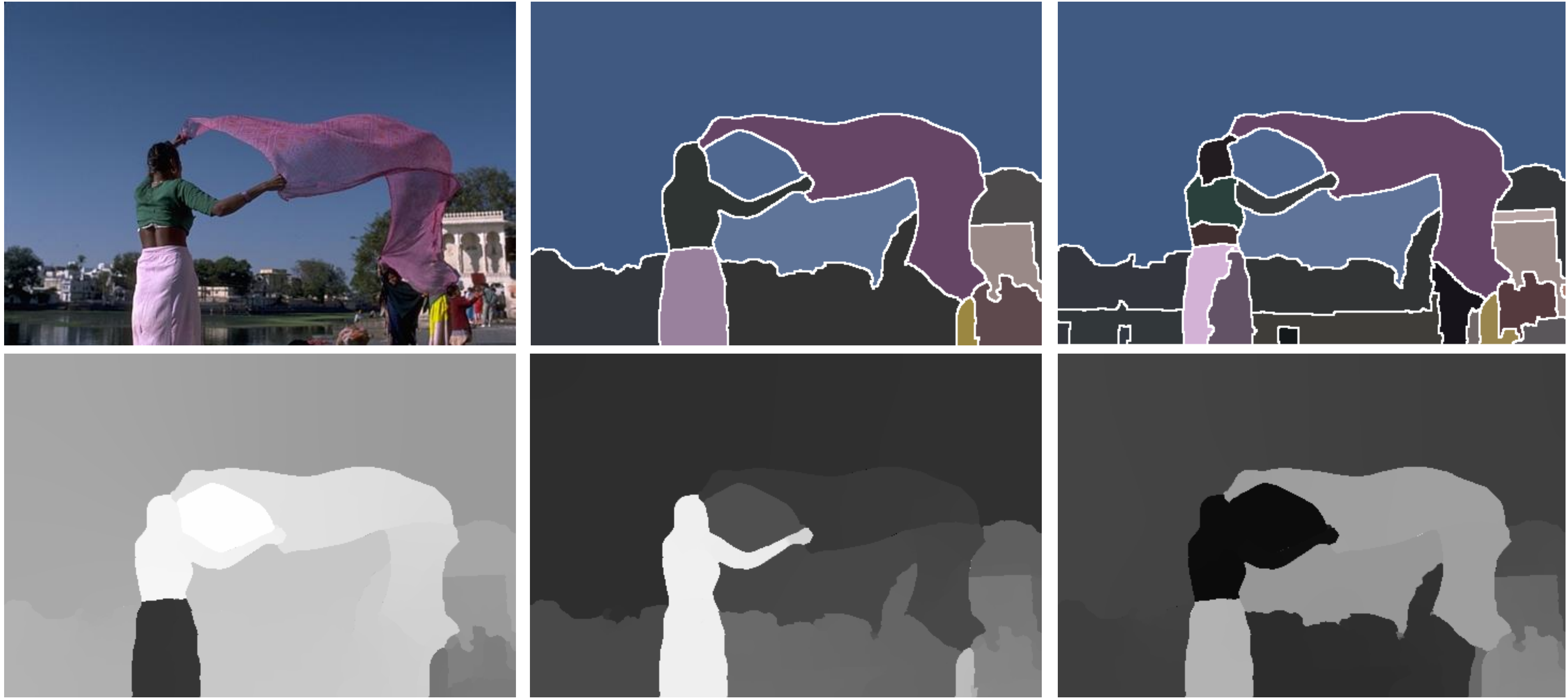}\vspace{-1mm}
    \caption{Embedding and segmentation results of piecewise flat embedding (PFE). Given an input image (top left), PFE transforms it into a set of embedding channels (bottom), each of which focuses on a subset of characteristics of the original image. A main highlight of our embedding is that the embedding channels tend to be piecewise flat. Two frameworks are tested to segment the image using embedding results, including clustering based segmentation~\cite{malik2001contour} (top middle) and contour driven hierarchical segmentation (top right)~\cite{arbelaez2011contour}.
    \vspace{-2mm}}
    \label{fig:teaser}
\end{figure}

A common practice to tackle this challenge embeds pixels into a new feature space by solving an optimization problem which is usually a continuous approximation of a balanced graph cut ~\cite{chan1994spectral,shi2000normalized,buhler2009spectral}. In this new feature space pixels with similar attributes are pulled closer while pixels with dissimilar attributes stay away from each other. Clustering pixels in this feature space would improve segmentation performance if distances in this space preserve intrinsic visual differences while being insensitive to distractions caused by shading and illumination. Nonetheless, because pairwise affinity is measured coarsely, $L_2$-regularized graph embedding method has noticeable smooth variations within object regions, which often make clustering-based segmentation algorithms produce false region boundaries~\cite{arbelaez2011contour}. One effective approach for removing such smooth variations computes nonlinear embeddings under various norms. $L_p$ ($1\leq p<2$) norm has been adopted in spectral clustering to produce 1-dimensional embedding with relatively small variations within clusters~\cite{buhler2009spectral,hein2010inverse}.

In this paper, we propose a new multi-dimensional nonlinear embedding, called \emph{piecewise flat embedding} (PFE), for image segmentation. This embedding is capable of suppressing variations within segments or clusters while preserving discontinuities between them. In the context of image segmentation, since object boundaries only occupy a small percentage of pixels in the whole image, in a locally connected graph, there only exist a sparse subset of pixel pairs whose pixels come from different object regions. 
An embedding to pursue the aforementioned property should only have a small subset of pixel pairs maintaining sufficiently large distances while distance between any remaining pixel pairs is pushed to zero.
Under the condition of locally connected graph with noisy similarity measurements, this embedding problem can be solved via minimizing an objective function regularized with the sparsity of nonzero pairwise pixel differences according to the theory of sparse signal recovery~\cite{donoho1989uncertainty,donoho1992signal}. Specifically, we attempt to promote sparse solutions with multiple dimensions by adopting an $L_{1,p}$-regularized ($0<p\leq 1$) energy term in the objective function.
This $L_{1,p}$-regularized embedding is a generalization of an $L_1$-regularized formulation, with extra space for sparsity pursuit and increased degree of nonconvexity as $p$ becomes smaller. 
A numerical solution is first developed to solve the $L_{1,1}$-regularized objective function by nesting two existing numerical solvers~\cite{goldstein2009split,lai2014splitting}. Orthogonally constrained embedding channels are simultaneously computed as local optima of the objective function. Inspired by iteratively reweighted $L_1$ algorithms (IRL1)~\cite{ochs2015iteratively}, we further design an iteratively reweighted algorithm based on the numerical solution to the $L_{1,1}$-regularized optimization to solve the more generic $L_{1,p}$-regularized problem.  

It is easy to integrate piecewise flat embedding into existing image segmentation frameworks, including clustering based image segmentation~\cite{malik2001contour} and contour-driven hierarchical segmentation~\cite{arbelaez2011contour}. An example of our embedding results and subsequent clustering based segmentation is shown in Figure~\ref{fig:teaser}. We conduct experiment on four popular benchmark datasets, including BSDS500~\cite{arbelaez2011contour}, Stanford Background Dataset (SBD)~\cite{gould2009decomposing}, the precise version~\cite{malisiewicz2007improving} of 23-class MSRC~\cite{shotton2006textonboost} and PASCAL Context~\cite{mottaghi2014role}. Both image segmentation frameworks incorporating PFE achieve state-of-the-art performance comparing to their counterparts.

This paper is an extended version of \cite{yu2015piecewise}. New technical contents in this version are summarized as follows.

{\noindent $\bullet$} The previous $L_{1,1}$-regularized objective function is extended to an $L_{1,p}$-regularized formulation, which still makes the embedding piecewise flat. An iteratively reweighted numerical algorithm based on nested Bregman iterations is devised to minimize this new objective function.

{\noindent $\bullet$} The overall numerical solution becomes an order of magnitude faster due to the adoption of
the Cholesky Decomposition~\cite{chen2008algorithm} for directly solving the sparse least-squares problems in the numerical pipeline.

{\noindent $\bullet$} New initialization schemes and a weighting strategy based on the residual cost of individual embedding channels are devised for piecewise flat embedding to achieve improved performance in image segmentation. 

{\noindent $\bullet$} Segmentation algorithms integrating our embedding produce state-of-the-art performance on four benchmark datasets (BSDS500, SBD, MSRC and PASCAL Context).

\section{Related Work}\label{sec:related-work}
\noindent\textbf{Image Segmentation:} A comprehensive summarization on segmentation methods has been given in  \cite{fan2001automatic}. Here we discuss 3 types of segmentation techniques which are closely related to our method. If pixels in an image are taken as a set of unordered data samples, most clustering algorithms, such as K-means, Gaussian Mixture Models~\cite{mclachlan2004finite} and MeanShift~\cite{comaniciu2002mean}, can be used for image segmentation.

\textbf{Gradient based methods} form the other representative approach in image segmentation. Examples include segmentation based on the watershed transform~\cite{najman1996geodesic}. In \cite{arbelaez2009contours}, OWT-UCM transforms a boundary probability map to a hierarchical segmentation, where oriented watershed transform (OWT) is designed to remove noises arising from the standard watershed algorithm, and ultrametric contour map (UCM) combines closed, non-self-intersecting weighted contour information. With the recent developments in boundary detection~\cite{martin2004learning,dollar2015fast,kokkinos2015pushing,Man+17}, OWT-UCM remains a favorable choice for image segmentation~\cite{arbelaez2011contour,pont2017multiscale,Man+17}. Starting from small superpixels, INSCRA~\cite{ren2013image} generates hierarchical segmentation with the help of a sequence of cascaded discriminative classifiers. LEP~\cite{zhaosegmenting} is a more recent hierarchical segmentation algorithm, which is inspired by the least effort principle and models the amount of effort involved in tracing boundaries by human.

\textbf{Graph partition} is another widely used approach. Modeling the 2D pixel grid of an image as a Markov random field, graphcut algorithms~\cite{greig1989exact,kolmogorov2004energy} have been developed to solve binary or multi-label image segmentation efficiently. Graph cuts~\cite{kolmogorov2004energy} solves the partition problem using an energy function consisting of a posterior term and a Potts model. The $L_1$ distance is adopted in the Potts model to preserve discontinuity. Graph based segmentation~\cite{felzenszwalb2004efficient} is another efficient method based on greedily deciding the existence of boundaries between pairs of regions. Spectral embedding methods~\cite{hagen1992new,chan1994spectral,shi2000normalized} project pixels into a new feature space where similar pairs of pixels stay closer than dissimilar pairs. The classical Ratio Cut problem~\cite{wei1989towards} is considered in \cite{hagen1992new,chan1994spectral} while Normalized Cut is put forward in~\cite{shi2000normalized}. Channels in a spectral embedding provide not only better features for clustering based segmentation~\cite{malik2001contour}, but also clearer global boundary information~\cite{maire2008using}. Cheeger Cut~\cite{cheeger1970lower} is able to set a lower bound of the second smallest eigenvalue of the graph Laplacian. B{\"u}hler and Hein~\cite{buhler2009spectral} utilize a uniform $p$-Laplacian formulation to define spectral embedding methods with various norms and prove that the optimal solution of Cheeger Cut can be found by thresholding the second eigenvector of the $p$-Laplacian when $p\rightarrow 1$. They have also devised an algorithm based on gradient and Newton steps to compute the second eigenvector for $1<p<2$. Their algorithm can be applied to divisive clustering/segmentation. The work in \cite{szlam2010total} and \cite{hein2010inverse} is dedicated to solving the second eigenvector of 1-Laplacian on the basis of Split Bregman and inverse power algorithms respectively. Our method is similar to $p$-Laplacian as we also adopt a variational objective function with sparsity regularization. However our method differs from them in two aspects. First, we seek to solve an $L_{1,p}$-regularized objective function where $0<p\leq1$. Second, spectral methods using $p$-Laplacian ($1\leq p<2$) only compute a single embedding channel every time, namely the second eigenvector, making them limited for recursive top-down clustering. Our objective function inherits the orthogonality constraint from a generalized version of normalized 2-Laplacian namely Laplacian Eigenmaps~\cite{belkin2003laplacian}, which enables multiple diverse embedding channels to be computed simultaneously. Thus our method can be conveniently integrated into both clustering and gradient based segmentation algorithms.

\vspace{1.5mm}

\noindent\textbf{Manifold Learning:} To cope with the "curse of dimensionality", various manifold learning algorithms have been proposed for nonlinear dimension reduction preserving structural information.
These algorithms roughly fall into two categories, local and global methods. Local methods, such as LLE~\cite{roweis2000nonlinear} and Laplacian Eigenmaps~\cite{belkin2003laplacian}, attempt to preserve local structures among data points. 
t-SNE~\cite{van2008visualizing}, an extension of SNE~\cite{hinton2002stochastic}, generates pleasant embedding results for the visualization of high dimensional data.
GNMF~\cite{cai2011graph} is another local method that proposes a graph regularized objective function to take into account both the reconstruction loss and local pairwise variation cost.

Global methods, such as Isomap~\cite{tenenbaum2000global} and structure preserving embedding~\cite{shaw2009structure}, attempt to preserve local and global relationships among all data points. Sparsity constraints have been considered in SMCE~\cite{elhamifar2011sparse} and Sparse Embedding~\cite{nguyen2012sparse}. SMCE constructs neighborhood relationships
via selecting neighbors from the same manifold for each data point, which is solved as a sparse optimization problem. It then follows the spectral method to compute embedding and clustering results. Sparse Embedding performs simultaneous dimension reduction and dictionary learning by learning a transformation from the original high-dimensional space to a low-dimensional space while preserving sparse structures. SSC proposed in~\cite{elhamifar2013sparse} attempts to learn a sparse and global affinity matrix through representing every sample with a few samples in the same subspace. The more the reference sample contributes to reconstructing the query sample, the higher similarity they possess. GraphEncoder~\cite{tian2014learning} tries to embed affinity matrix into feature space directly in the prototype of autoencder. \cite{chang2015heterogeneous} proposes to embed different types of data (image and text) to a unified space with the help of deep learning technologies. Our method differs from these embedding techniques in the sense that we directly minimize a global objective function formulated using $L_{1,p}$ regularization without dictionary learning and achieves high performance in the image segmentation task.

\begin{figure}[!t]
    \centering
    \includegraphics[width=\columnwidth]{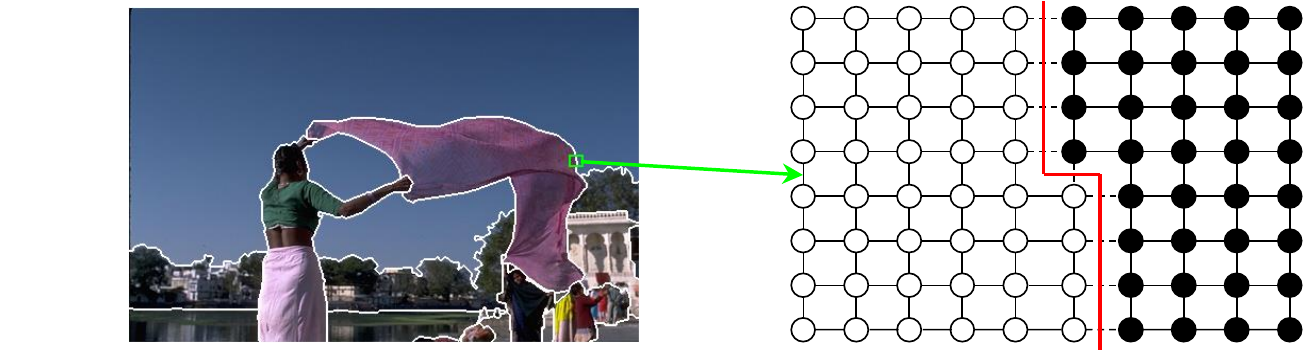}\\
    \includegraphics[width=\columnwidth]{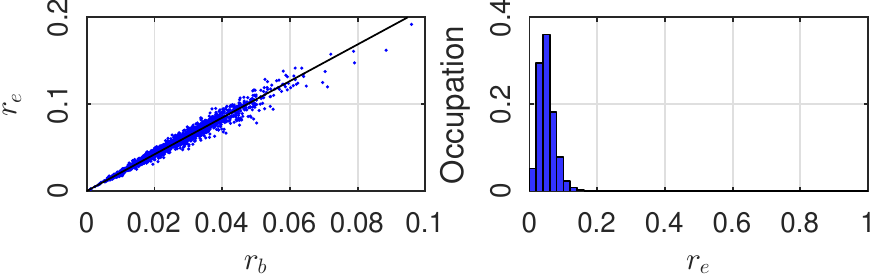}%
    \vspace{-3mm}
    \caption{Sparsity of edges crossing region boundaries in a locally connected graph. The ratio $r_b$ of boundary length against whole image area in the top-left image is 2.409\%. The $\mathcal{N}_4$-connected graph of a $8\times10$ patch is visualized (top-right). It is divided by a boundary shown in red. Only 9 among 142 edges cross the boundary. Actually the percentage of such edges $r_e$ depends on the length and shape of boundaries. Statical observations in all 2697 groundtruth segmentation maps of BSDS500 are provided in bottom row: approximately linear relation between $r_e$ and $r_b$ (left); distribution of $r_e$ among these segmentation maps (right). Here every pixel is connected to others in a 11$\times$11 neighborhood. Average of $r_b$ and $r_e$ is 2.413\% and 5.113\% respectively. }\vspace{-4mm}  
    \label{fig:sparsity}
\end{figure}

\section{Piecewise Flat Embedding}\label{sec:pfe}
\noindent Let us consider the problem of dividing an image into regions containing pixels with similar low-level/high-level features. We first construct a locally connected graph $G=(V,E)$, where $V$ represents the set of pixels and $E$ represents edges connecting pixels within their neighborhoods. Usually there is only a very small fraction of edges which straddle across region boundaries because of the sparseness of boundaries in 2D image plane -- the majority are between pixels from the same region.
Intuitions and statistical observations in BSDS500 are provided in Figure~\ref{fig:sparsity}. These indicate region boundaries might be discovered as the solution to an energy minimization problem with sparsity regularization.

Further, we would like to constrain the feature embedding of pixels from the same region to lie as tightly with one another as possible, therefore achieving the goal of piecewise ``flat'' embedding. This piecewise flatness property matters. From one point of view it reduces the intra-class variations of raw pixel values within image segments, for reasons explained earlier. From another point of view, it aims to factorize out a \emph{piecewise smooth} component, which is known as an intrinsic component of pixel values from image formation, and pursues a feature embedding that captures a set of \emph{piecewise constant} representations. This section will describe our formulation of the problem and a numerical solution to it.



\subsection{Problem Definition} \label{sec:pfedef}
\noindent Given $n$ data points $\mathbf{X} =\{\boldsymbol x_1,\cdot\cdot\cdot, \boldsymbol x_n\}$ in $R^m$, we would like to embed them into a new $d$-demension space $R^d$. Let $\mathbf{Y}=[\mathbf{y}_1,\cdot\cdot\cdot,\textbf{y}_n]^\text{T}$  be the $n$$\times$$d$ data matrix in the new space, where $\mathbf{y}_k^\text{T}$ is the $k$-th row of $\mathbf{Y}$, representing the $d$ coordinates of the $k$-th data point in the new space. 
PFE is to find an embedding $\mathbf{Y}$ that achieves the sparsity, or the piecewise flat property by minimizing the following objective function,
\begin{equation}\label{eq:PFEobj}
    \min_{\mathbf{Y}}\sum_{ij}{\left\|W_{ij}(\mathbf{y}_{i} - \mathbf{y}_{j})\right\|_{1}^{p}} \hspace{5mm} s.t. \hspace{2mm}\mathbf{Y}^{\text T}\mathbf{DY} = \mathbf{I},
\end{equation}
where $0<p\le 1$, $\mathbf{W}=\{W_{ij}\}$ is the symmetric affinity matrix, and $\mathbf{D}$ is the diagonal weight matrix with $D_{ii} = \sum_{j}W_{ji}$. Our embedding formulation is inspired by Normalized Cut and Laplacian Eigenmaps (LE)~\cite{belkin2003laplacian}, which is defined as follows:
\begin{equation}\label{eq:LEobj}
    \min_{\mathbf{Y}}\sum_{ij}{W_{ij}\left\|\mathbf{y}_{i} - \mathbf{y}_{j}\right\|_2^2} \hspace{5mm} s.t. \hspace{2mm}\mathbf{Y}^{\text T}\mathbf{DY} = \mathbf{I},
\end{equation}
where $\mathbf{W}$ and $\mathbf{D}$ are defined in the same way.

Note $\mathbf{W}$ is a symmetric sparse matrix, 
so we can rewrite the problem formulation in (\ref{eq:PFEobj}) as follows,
\begin{equation}\label{eq:PFEobjL1p}
    \min_{\mathbf{Y}}{\left\|(\mathbf{M}\mathbf{Y})^{\text{T}}\right\|_{1,p}^{p}} \hspace{5mm} s.t. \hspace{2mm}\mathbf{Y}^{\text T}\mathbf{DY} = \mathbf{I},
\end{equation}
where $\mathbf{M}=\{M_{ij}\}$ is a $t$$\times$$n$ matrix, $M_{ki}=W_{ij}=-M_{kj}$ are for data points $\boldsymbol x_i$ and $\boldsymbol x_j$ that form the $k$-th non-zero entry of the affinity matrix, and $t$ is the total number of non-zero entries. $\left\|\mathbf{A}\right\|_{q,p} = (\sum_{j}(\sum_{i}{|A_{ij}|^q})^{p/q})^{1/p}$ is the $L_{q,p}$ norm of matrix $\mathbf{A}$.

Similarly, the Laplacian Eigenmaps formulation (\ref{eq:LEobj}) can be rewritten as follows,
\begin{equation}\label{eq:LEobjs}
    \min_{\mathbf{Y}}{\left\|(\mathbf{M^\prime}\mathbf{Y})^{\text{T}}\right\|_{2,2}^{2}} \hspace{5mm} s.t. \hspace{2mm}\mathbf{Y}^{\text T}\mathbf{DY} = \mathbf{I},
\end{equation}
where $\mathbf{M^\prime}$ is a $t$$\times$$n$ matrix, $M_{ki}^{\prime}=W_{ij}^{1/2} = -M_{kj}^{\prime}$ are for data points $\boldsymbol x_i$ and $\boldsymbol x_j$ that form the $k$-th non-zero entry of $\mathbf{W}$.

Although (\ref{eq:PFEobjL1p}) and (\ref{eq:LEobjs}) are similar,
the most obvious difference is that we use the $L_{1,p}$ norm in (\ref{eq:PFEobjL1p}) while the formulation of Laplacian Eigenmaps inherits a common trait of mainstream multi-dimensional embedding methods, which is the use of the squared $L_{2,2}$ norm in the objective function.
In fact, such a small difference has the following important implication.
The $L_{1,p}$ norm promotes sparse solutions while the $L_{2,2}$ norm does not. A sparse solution implies that, among all locally connected point pairs, there only exists a sparse subset of point pairs whose distances in the new space are sufficiently large, and the distances between the rest of the point pairs are almost zero. Such a sparse solution suggests points with a large distance in the new space should belong to different clusters and points with a very small distance in the new space should belong to the same cluster. Therefore, performing clustering in the new space becomes a straightforward process. On the other hand, although the affinity matrix in (\ref{eq:LEobjs}) attempts to move points with similar attributes closer in the new space, the resulting pairwise distances in the new space still follows a relatively smooth distribution because the measurement of pairwise affinity is coarse and noisy, and the distinction between similar points and dissimilar points is not as clear as in our results.

In the context of image segmentation, a sparse solution to (\ref{eq:PFEobjL1p}) means there only exist a sparse subset of locally connected pixel pairs whose pairwise difference in the new space, $\mathbf{y}_{i}-\mathbf{y}_{j}$, remains sufficiently different from zero. Since the affinity $W_{ij}$ appears together with $\|\mathbf{y}_{i}-\mathbf{y}_{j}\|_1$, a large affinity forces $\|\mathbf{y}_{i}-\mathbf{y}_{j}\|_1$ to approach zero and $\|\mathbf{y}_{i}-\mathbf{y}_{j}\|_1$ becomes nonzero only when the affinity is relatively small. On the other hand, pixel pairs crossing region boundaries typically have smaller affinities than those from the same region. Therefore, the sparse solution to (\ref{eq:PFEobjL1p}) should be capable of facilitating the discovery of region boundaries.
At the interior of a region, $\mathbf{y}_{i}-\mathbf{y}_{j}$ would be most likely zero due to the large affinity between them. This implies each embedding channel has an almost constant value inside every region, hence, the name {\em piecewise flat embedding}.

\begin{figure}[t]
    \centering
    \includegraphics[width=\columnwidth,clip,trim=0 0 0 400]{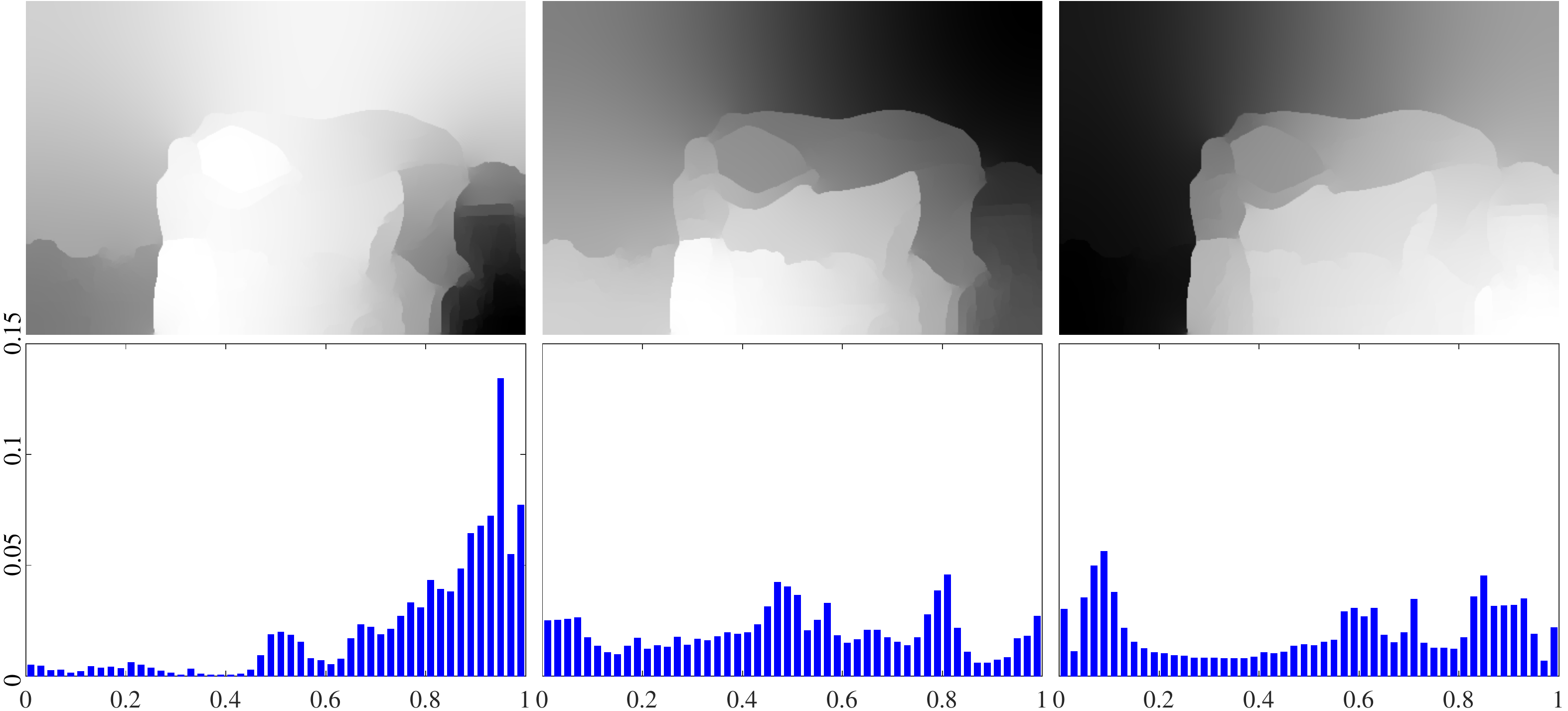}
    \includegraphics[width=\columnwidth,clip,trim=0 412 0 0]{embeddinghists7.pdf}\vspace{-2mm}
    \caption{Laplacian Eigenmaps (top row) and PFE (third row) embedding results, and histograms of each embedding channel. PFE channels are almost piecewise constant while LE channels are piecewise smooth. Because of the adopted $L_{1,p}$ norm, the histograms of our results are much sparser than those of Laplacian Eigenmaps. }\vspace{-4mm}
    \label{fig:embhists}
\end{figure}

A visual comparison between our $L_{1,1}$-regularized PFE and Laplacian Eigenmaps are shown in Figure~\ref{fig:embhists}, where each image represents pixelwise values of one embedding channel and the histogram below each image represents the pixel distribution of that embedding channel. According to the results in this figure, PFE channels are almost piecewise constant and their histograms are much sparser than those of Laplacian Eigenmaps, which still possess obvious nonzero gradients within image regions.

\noindent \textbf{Constraint and Total Variation}
The constraint in (\ref{eq:PFEobjL1p}), which is inherited from Laplacian Eigenmaps, can be decomposed into the following ones,
\begin{align}
\label{eq:normal} \min_{\mathbf{Y}}{\left\|(\mathbf{M}\mathbf{Y})^{\text{T}}\right\|_{1,p}^{p}} \hspace{4mm} s.t. \; &\forall u,\; y_u^{\text{T}}\mathbf{D}y_u=1;  \\
\label{eq:orthog} &\forall u,\; \forall v\neq u,\;  y_u^{\text T}\mathbf{D}y_v=0 ;
\end{align}
where $y_u$ is the $u$-th column of $\mathbf Y$. The first condition (\ref{eq:normal}) indicates every embedding channel is balanced using a normalized criterion as in~\cite{shi2000normalized}. The second condition (\ref{eq:orthog}) makes different embedding channels orthogonal to each other, which improves the diversity of the embedding channels.


The discretized total variation regularization~\cite{rudin1992nonlinear} has been frequently used in computer vision tasks such as image smoothing~\cite{min2014fast,BiHY15} and motion estimation~\cite{werlberger2010motion}. A typical image restoration/smoothing objective function with weighted nonlocal total variation regularization~\cite{gilboa2008nonlocal} can be written as follows,
\begin{equation}
\min_{\mathbf{Y}} \| \mathbf{Y}-\mathbf{Y}_0\|_{2,2}^2+\kappa\sum_{ij}W_{ij}\|\mathbf y_i-\mathbf y_j\|_2^2,
\label{eq:WNonlocalTV}
\end{equation}
where $\mathbf Y$ represents the restored/smoothed result and $\mathbf Y_0$ is the input image. $\kappa$ is a constant and $W_{ij}$ is usually computed from a guidance image. The first term in the above formulation is a data term, which uses the input image as a soft constraint to avoid a trivial optimal solution. The above problem can be solved using weighted least squares~\cite{farbman2008edge}. $L_p$-regularized total variation has been used in \cite{min2014fast,BiHY15}. Dynamic guidance has been exploited with the help of Welsch's function in~\cite{ham2015robust}. Because of the data term, results from total variation formulations are typically enhanced/restored images whose channels strictly correspond to channels in the input image. Our objective function in (\ref{eq:PFEobj}) utilizes a similar regularization with the $L_{1,p}$ norm. However, the data term in total variation formulations is replaced with the orthogonality constraint in our formulation. The adoption of the orthogonality constraint gives us more freedom to define novel embedding spaces that are more independent of the input image. For instance, the dimensionality of an embedding space can be different from that of the input image, and embedding channels do not have to correspond to channels of the input image. Nonetheless, this freedom brings numerical challenges.

\subsection{Numerical Solution} \label{sec:algorithm}
\noindent In the following, we first present an iterative numerical solution for a special case of the problem in (\ref{eq:PFEobjL1p}), where the $L_{1,1}$ norm is used instead of the more general $L_{1,p}$ norm.
Then a numerical solution for the original problem in (\ref{eq:PFEobjL1p}) with the $L_{1,p}$ norm is introduced as an iteratively reweighted version of the special-case solution.


\subsubsection{Solution to ${L_{1,1}}$-regularized Objective Function}\label{sec:L11}
\noindent By setting $p=1$ in (\ref{eq:PFEobjL1p}), we obtain the ${L_{1,1}}$-regularized objective function as follows,
\begin{equation}\label{eq:PFEobjL11}
    \min_{\mathbf{Y}}{\left\|\mathbf{M}\mathbf{Y} \right\|_{1,1}} \hspace{5mm} s.t. \hspace{2mm}\mathbf{Y}^{\text T}\mathbf{DY} = \mathbf{I}.
\end{equation}
Solving this problem is challenging as it consists of $L_{1,1}$ regularization and an orthogonality constraint while most of existing numerical algorithms can only handle one of them. Here we present a numerical solution by nesting two existing solvers that handle $L_1$-regularized optimization and orthogonality constraints respectively. This solution is later accelerated with a two-stage strategy that relaxes the orthogonality constraint.

Since the objective function in (\ref{eq:PFEobjL11}) is convex, we apply the Splitting Orthogonality Constraint (SOC) algorithm in \cite{lai2014splitting}. Following the derivation of the SOC algorithm, we define $\mathbf{P} = \mathbf{D}^{1/2}\mathbf{Y}$, and rewrite (\ref{eq:PFEobjL11}) as,
\begin{equation}\label{eq:PFEobjL1d1}
    \min_{\mathbf{Y}}\, \|\mathbf M \mathbf Y \|_{1,1} \hspace{3mm} s.t. \hspace{2mm}\mathbf{D}^{1/2}\mathbf{Y} = \mathbf{P}, \hspace{1mm}\mathbf{P}^{\text T}\mathbf{P} = \mathbf{I},
\end{equation}
which can be iteratively solved using Bregman iterations~\cite{bregman1967relaxation} as follows:

\begin{itemize}
  \setlength\itemsep{1em}
  \item[(a)]
  $\mathbf{Y}^{(k+1)}  = \arg \underset{\mathbf{Y}} \min \;  \|\mathbf M \mathbf Y \|_{1,1} + \frac{r}{2}||\mathbf{D}^{1/2}\mathbf{Y} - \mathbf{P}^{(k)} + \mathbf{B}^{(k)} ||_{2,2}^2 $;
  \item[(b)]
    $\mathbf{P}^{(k+1)} = \arg \underset{\mathbf{P}} \min \; \| \mathbf{P} - (\mathbf{D}^{1/2}\mathbf{Y}^{(k+1)}+\mathbf{B}^{(k)}) \|_{2,2}^2$,\\
     $ \text{s.t.} \;\; \mathbf{P}^\text{T}\mathbf{P} = \mathbf{I}$;
  \item[(c)]
  $\mathbf{B}^{(k+1)} = \mathbf{B}^{(k)} + \mathbf{D}^{1/2}\mathbf{Y}^{(k+1)} - \mathbf{P}^{(k+1)}$;
\end{itemize}
where $\mathbf{B}$ is an auxiliary matrix, and $\mathbf{B}^{(0)}=\mathbf 0$.



According to Theorem 2.1 in \cite{lai2014splitting}, the spherically constrained problem in step (b) has the following closed form solution:
\begin{eqnarray}
  \nonumber
  \mathbf{D}^{1/2}\mathbf{Y}^{(k+1)} + \mathbf{B}^{(k)} &=& \mathbf{U}\mathbf{\Sigma}_{n\times d}\mathbf{V}^{\text T}, \\
  \mathbf{P}^{(k+1)} &=& \mathbf{U} \mathbf{I}_{n\times d}\mathbf{V}^\text{T}, \label{eq:ortho}
\end{eqnarray}
where $\mathbf{U}$ and $\mathbf{V}$ are matrices with orthogonal columns from the SVD decomposition in the first step.

The subproblem in step (a) of the above numerical solution inherits the $L_{1,1}$-regularized energy term from (\ref{eq:PFEobjL11}). There exist many numerical solutions for optimization problems with $L_1$ regularization. For reasons discussed later, we choose to apply the Split Bregman algorithm in \cite{goldstein2009split} to solve the subproblem in step (a). By introducing auxiliary variables, the Split Bregman algorithm solves an $L_1$-regularized optimization iteratively by transforming the original optimization into a series of differentiable unconstrained convex optimization problems. The definition of any convex optimization in the series depends on the auxiliary variables passed from the previous iteration, and convergence can be achieved within a relatively small number of iterations. Detailed solution to the $L_{1,1}$-regularized problem in step (a) is enclosed in Appendix~\ref{appendI}.

\vspace{2mm}
\noindent \textbf{Two-Stage Implementation}
In practice, we run the following two-stage implementation to obtain a high-quality solution more efficiently.

\noindent \emph{Stage I}
This stage implements the full numerical solution with nested Bregman iterations. A large penalty coefficient for the orthogonality constraint is used. The outer loop makes different dimensions of the embedded data orthogonal to each other to remove redundancy among them. This is important in avoiding naive solutions with highly redundant or even duplicate dimensions.

However, orthogonality is a highly non-convex constraint that prevents the objective function in (\ref{eq:PFEobjL11}) from settling into a truly low-energy state; and more important, it is not absolutely necessary for us to pursue an embedding whose dimensions are strictly orthogonal as long as there is not too much redundancy across different dimensions. Therefore, following a few iterations of the full numerical solution, we relax the orthogonality constraint in a second stage.

\noindent \emph{Stage II} This stage only executes the Bregman iterations in the inner loop without performing the SVD in the outer loop to strictly enforce orthogonality. Such relaxation of the orthogonality constraint allows the objective function to reach a lower energy.

\begin{figure}[t]
  \centering
  \includegraphics[width=0.8\columnwidth]{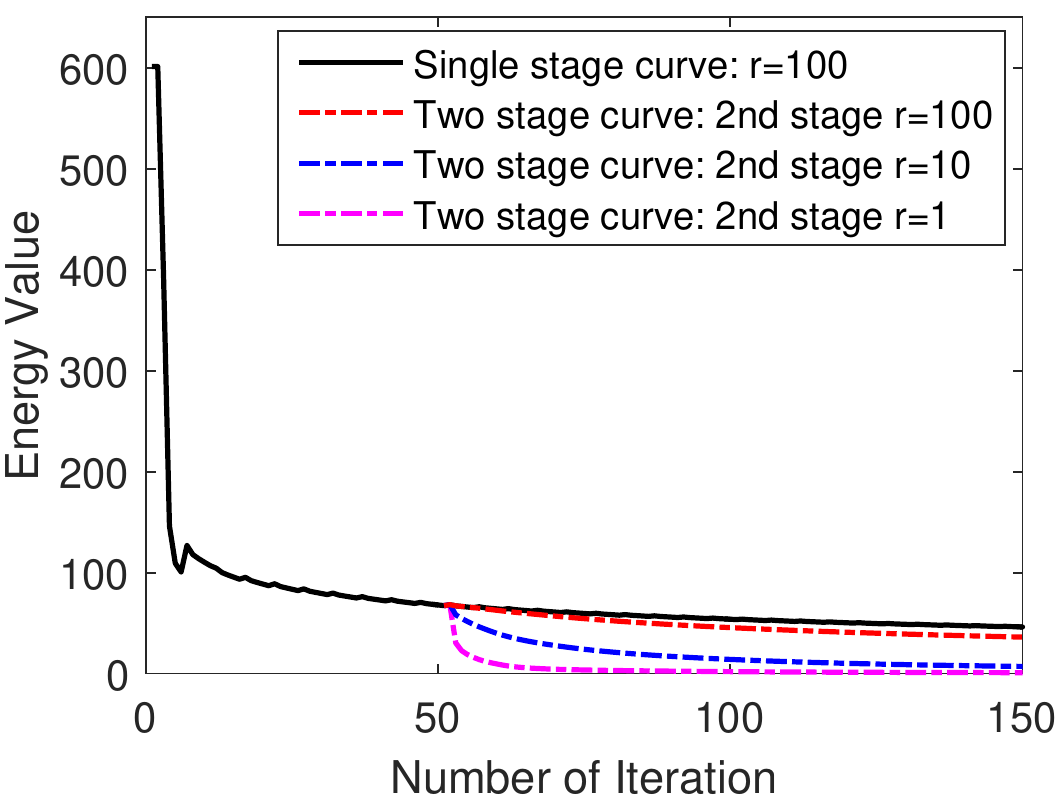}\vspace{-3mm}
  \caption{Energy curves. The black curve shows how the energy decreases with regular single-stage Bregman iterations with $\lambda = 1000, r = 100$. The dashed curves show how energy evolves with our 2-stage algorithm. The first stage is the same as the black curve. The second stage starts from iteration 50. The three dashed curves show varying convergence rates using different parameter values ($r = 100, 10, 1$ respectively for the red, blue and pink curves.)}\label{fig:energycurve}\vspace{-3mm}
\end{figure}

Figure \ref{fig:energycurve} shows an example of the energy curve during Stage I and Stage II. It verifies our two-stage scheme can reach a lower energy than a single-stage one.

\begin{figure}[t]
  \centering
  \includegraphics[width=\columnwidth]{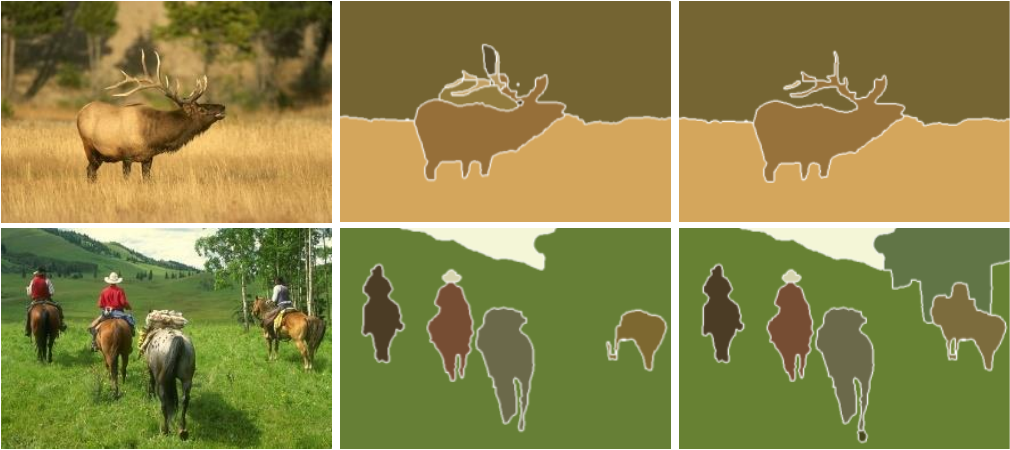}\vspace{-1mm}

  \caption{Comparison of segmentations using piecewise flat embedding without (middle) and with (right) residual-based channel weighting. Left are input images.}\label{fig:CMPOvsW}\vspace{-2mm}
\end{figure}

\vspace{1.5mm}
\noindent \textbf{Residual-Based Channel Weighting} For the sake of efficiency, we run the numerical procedure in each stage for a predefined number of iterations instead of waiting until the empirical error of every embedding channel falls below a predefined tolerance $\epsilon$.
Thus the obtained embedding channels differ from each other with respect to the degree of convergence. To moderate the effect of such early termination, we devise the following weighting scheme based on the residual cost of the objective function with respect to each embedding channel,
\begin{equation} \label{eq:weighting}
 \eta_{v}=\sqrt{\frac{\|\mathbf M y_{v}\|_1} { \|\mathbf D^{1/2}y_v\|_2}},\quad
\hat{y}_v=\mbox{normalize}(y_v)/\eta_{v},
\end{equation}
where $y_{v}$ and $\hat{y}_v$ are the $v$-th columns of $\mathbf Y$ obtained at the end of Stage II and the final embedding $\mathbf{\hat{Y}}$ respectively. This weighting scheme makes the importance of a channel lower if it has a larger residual cost. A comparison of PFE-based segmentations with and without residual-based channel weighting are shown in Figure \ref{fig:CMPOvsW}, where segmentation results with residual-based channel weighting are clearly better.

\begin{figure}[t]
    \centering
    \includegraphics[width=\columnwidth]{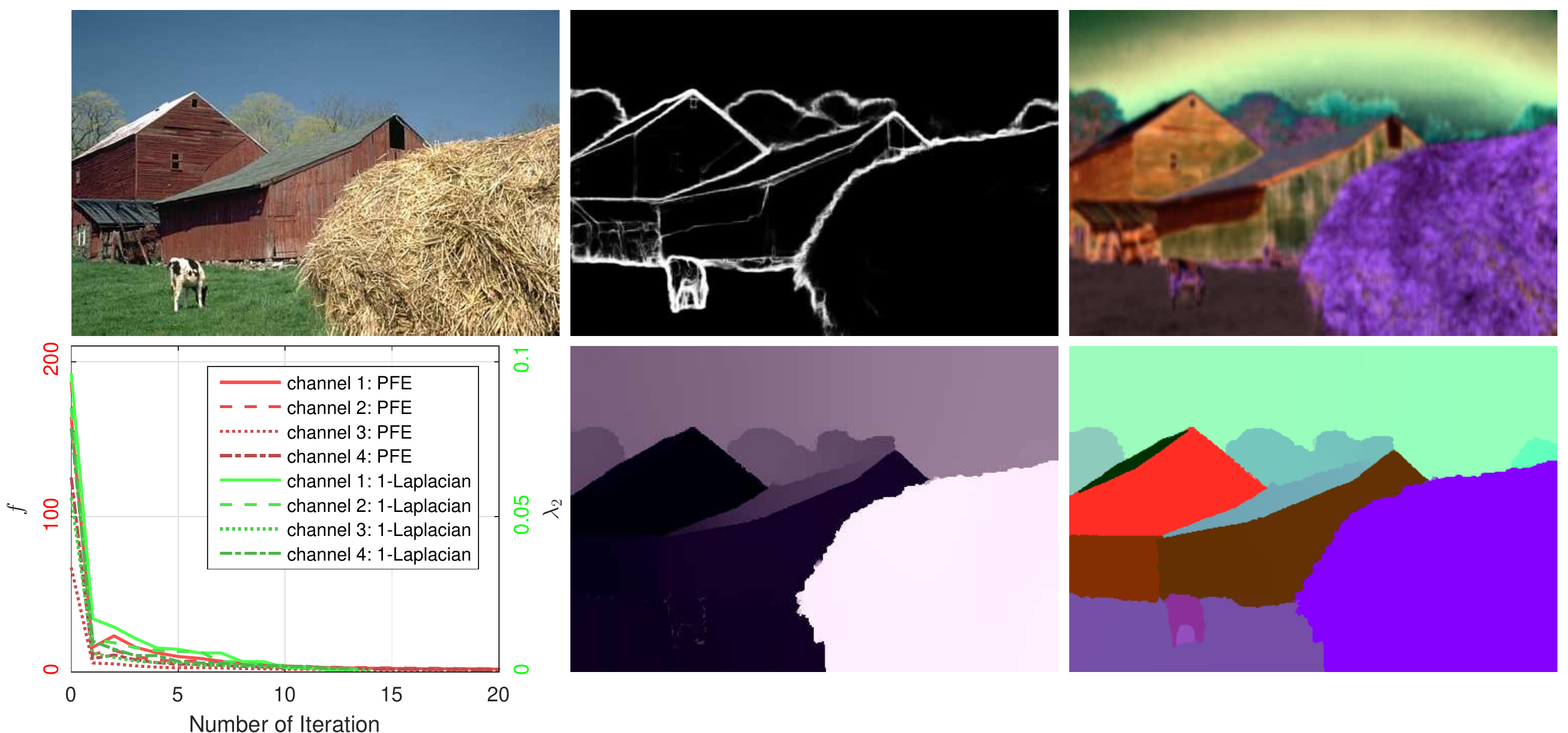}\vspace{-1mm}
    \caption{Comparison between PFE and 1-Laplacian~\cite{hein2010inverse}. Top row from left to right: input image; detected boundary map using~\cite{kokkinos2015pushing} for affinity matrix computation; initialization based on weighted spectral clustering as in Section~\ref{sec:ini}. Bottom row from left to right: curves of energy value $f$ (PFE) and the second smallest eigenvalue $\lambda_2$ (1-Laplacian); result of 1-Laplacian; result of PFE. 4 channels are computed in practice but only 3 embedding channels are visualized as color channels (best viewed in close-up and color image).}
    \label{fig:1laplacian}\vspace{-1mm}
\end{figure}
\vspace{1.5mm}
\noindent \textbf{Comparison to 1-Laplacian} The inverse power method in~\cite{hein2010inverse} is adopted to compute the eigenvector of 1-Laplacian with the second smallest eigenvalue. Our method enables multiple diverse embedding channels to be computed concurrently with the help of the orthogonality condition in (\ref{eq:orthog}). Comparisons of embedding results and convergence rates are shown in Figure~\ref{fig:1laplacian}. The same affinity matrix is computed from a boundary map~\cite{kokkinos2015pushing} for both methods. 1-Laplacian is applied to multiple distinct initializations independently to obtain a multi-channel embedding. The optimization processes of our method and 1-Laplacian are presented using energy value $f=\|\mathbf My\|_1$ and the second smallest eigenvalue $\lambda_2$ respectively. The convergence rates of our algorithm and 1-Laplacian are comparable. 1-Laplacian gives rise to similar embedding results across all channels despite different initializations while our method computes much more diverse channels. On average, it costs $87$s to compute one channel using 1-Laplacian and $58$s to compute 4 channels at once using our method.

\begin{figure}[t]
  \centering
  \includegraphics[width=\columnwidth,clip,trim=0 145 0 0]{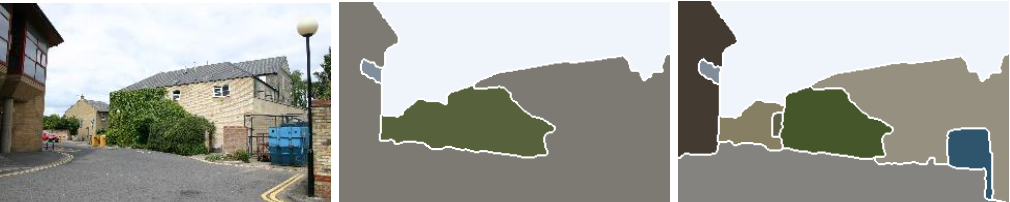}\vspace{.5mm}
  \includegraphics[width=\columnwidth,clip,trim=0 0 0 145]{CMP1vsP.pdf}\vspace{-1mm}
  \caption{Comparison of PFE-based segmentations with $L_{1,1}$ (middle) and $L_{1,0.8}$ (right) regularization. Left are input images.}\label{fig:CMP1vsP}\vspace{-2mm}
\end{figure}

\subsubsection{Solution to ${ L_{1,p}}$-regularized Objective Function}\label{sec:L1p}
\noindent To leverage algorithms for handling simpler objective functions when a more sophisticated objective function is optimized, a class of numerical methods based on iterative reweighting has been previously developed. These methods iteratively optimize a smooth or nonsmooth convex objective function dynamically weighted by coefficients evaluated from the previous iteration. For instance, IRLS\cite{daubechies2010iteratively} solves iteratively reweighted quadratic problems while IRL1~\cite{candes2008enhancing} solves iteratively reweighted $L_1$-regularized problems. Algorithms using iterative reweighting have been recently summarized in \cite{ochs2015iteratively}.
Inspired by such methods, we develop a method based on iterative reweighting to solve the $L_{1,p}$-regularized problem in (\ref{eq:PFEobjL1p}).

As done for the $L_{1,1}$-regularized problem in (\ref{eq:PFEobjL11}), the $L_{1,p}$-regularized problem in (\ref{eq:PFEobjL1p}) is also solved in two stages. The first stage of the numerical procedures for solving these two problems is completely the same. The result from the first stage initializes the second stage. In the second stage, we revise the objective function in step (a) of the procedure solving (\ref{eq:PFEobjL11}) as follows,
\begin{equation}\label{eq:PFEoptL1p}
\mathbf Y = \arg \min_{\mathbf Y} \|(\mathbf M \mathbf Y)^{\text{T}}\|_{1,p}^{p}+ \frac{r}{2}\Psi(\mathbf Y),
\end{equation}
where $\Psi(\mathbf Y) = \|\mathbf D^{1/2} \mathbf Y - \mathbf P + \mathbf B \|_{2,2}^{2}$, $\mathbf P$ and $\mathbf B$ are obtained from the last outer iteration of the first stage.

We apply the majorization-minimization algorithm~\cite{ochs2015iteratively} to solve the problem in (\ref{eq:PFEoptL1p}). A Detail description is enclosed in Appendix~\ref{appendII}.

Residual-based weighting introduced earlier can still be applied to individual channels of $\mathbf Y^{(k+1)}$ obtained in the first step to improve segmentation performance. Define the result from residual-based weighting as $\mathbf{\hat{Y}}^{(k+1)}$. Then the second step of the algorithm in Appendix~\ref{appendII} is revised as follows,
\begin{equation}\label{eq:reweighting}
w_{i}^{k+1} = \mbox{max}(\alpha\|\mathbf m_i^{\text{T}} \mathbf{\hat{Y}}^{(k+1)} \|_{1},\varepsilon)^{p-1},
\end{equation}
where $\alpha$ is a constant. We experimentally found that the revised iterations can converge when $0<\alpha\le 50$. 

A comparison of segmentations obtained through $L_{1,1}$ regularization and $L_{1,p}$ ($p=0.8$) regularization are shown in Figure \ref{fig:CMP1vsP}. $L_{1, 0.8}$ regularization gives rise to more accurate results.

\begin{figure}[t]
\centering
\includegraphics[width=\linewidth]{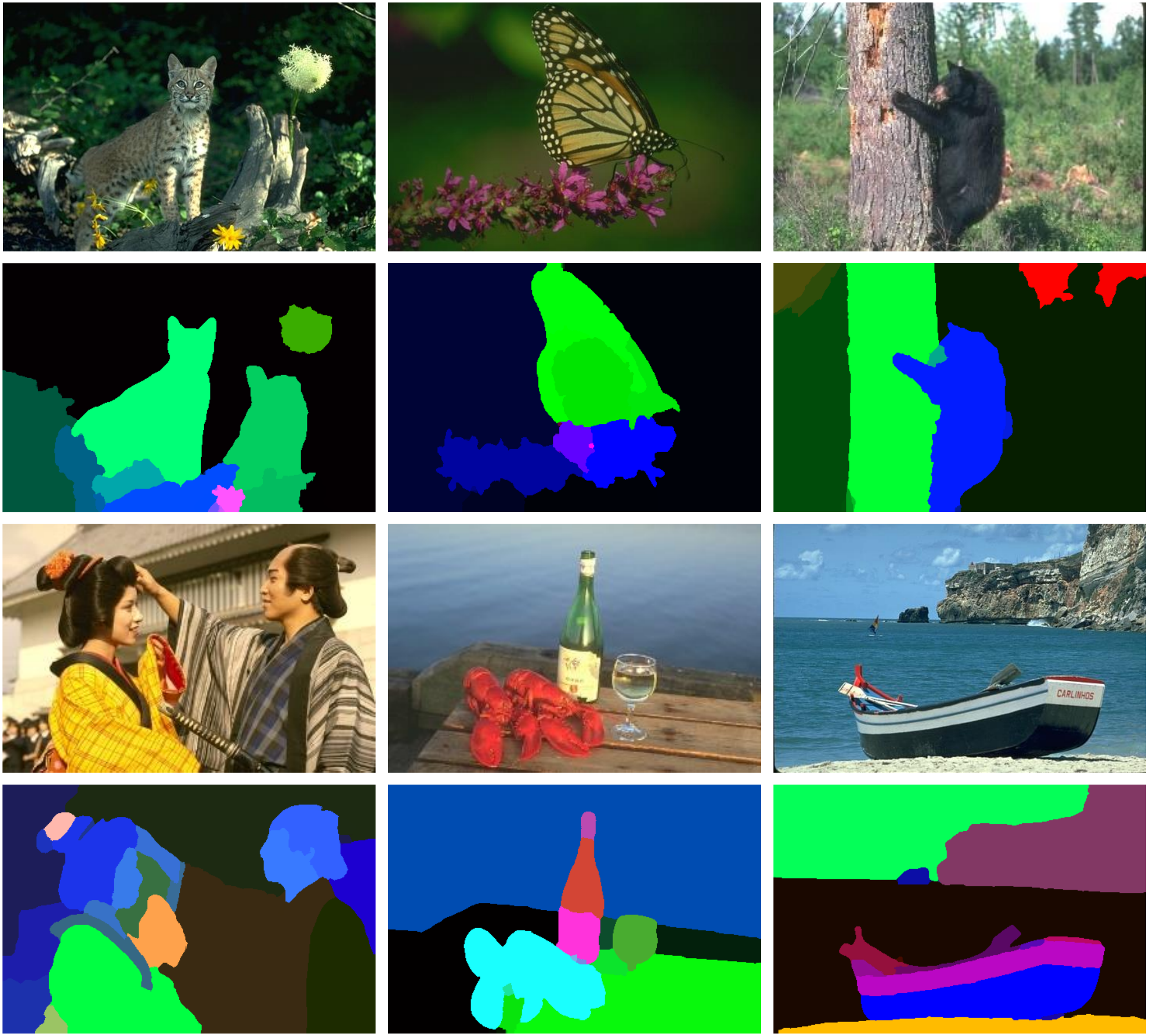}\vspace{-1mm}
\caption{
Examples of our piecewise flat embedding. Input images are located in the first and third rows. Images in the second and fourth rows are generated by visualizing 3 embedding channels as color channels. Note that the embedding results clearly exhibit piecewise flatness.}
\label{fig:embedding}\vspace{0mm}
\end{figure}

\noindent \textbf{Parameter Setting} For methods developed in this paper, we select parameters as follows. In Stage I, we set the maximum number of outer iterations to 5 and the number of inner iterations to 8. In Stage II of the algorithm optimizing the $L_{1,1}$-regularized objective function in (\ref{eq:PFEobjL11}), the maximum number of iterations is set to 40. For the $L_{1,p}$-regularized objective function in (\ref{eq:PFEobjL1p}), there still exists a double loop in the second stage due to iterative reweighting. We set the maximum number of outer iterations to 5 and the number of inner iterations to 20 or less.
The parameter setting for $\alpha$, $\varepsilon$, $\lambda$ and $r$ will be given in the following section on image segmentation. Figure \ref{fig:embedding} shows a few embedding results. 

\section{Image Segmentation Using Piecewise Flat Embedding}\label{sec:segmentation}
\begin{figure*}[t]
\centering
\includegraphics[width=\linewidth,clip,trim=0 10 0 10]{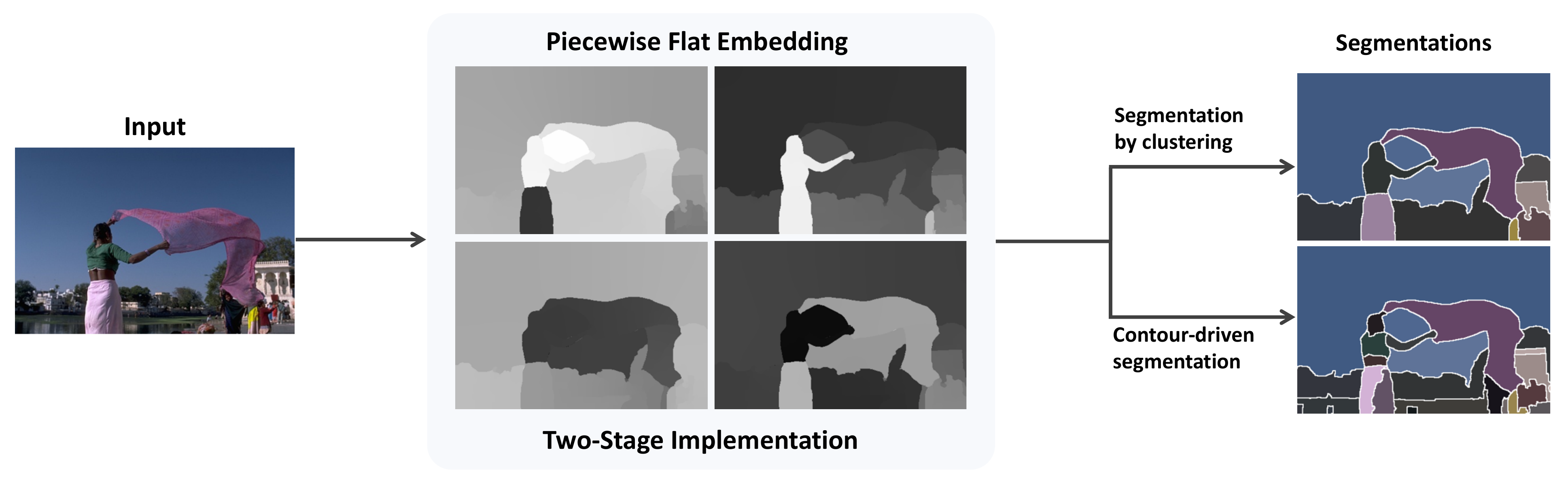}\vspace{-2mm} 
\caption{
Image segmentation pipeline. Given an input image, our method generates a piecewise flat embedding from an affinity matrix of the image in a two-stage optimization. The channels in the embedding are used for image segmentation either in a clustering-based method, or a contour-driven hierarchical method.}\vspace{0mm}
\label{fig:pipeline}
\end{figure*}

\noindent An image with resolution $w\times h$ gives rise to a set of $n = w\times h$ data points. By computing an affinity value between neighboring pixels, we have a locally connected graph (with $n$ nodes), where edge weights can be represented with a $n\times n$ sparse affinity matrix $\mathbf{W}$. A piecewise flat embedding of these data points in a new $d$-dimensional space can be computed using the numerical algorithms in the previous section. We have integrated such embedding results in two popular image segmentation frameworks.
Figure~\ref{fig:pipeline} illustrates our segmentation pipeline.
\begin{figure*}[t]
\centering
\includegraphics[width=\linewidth]{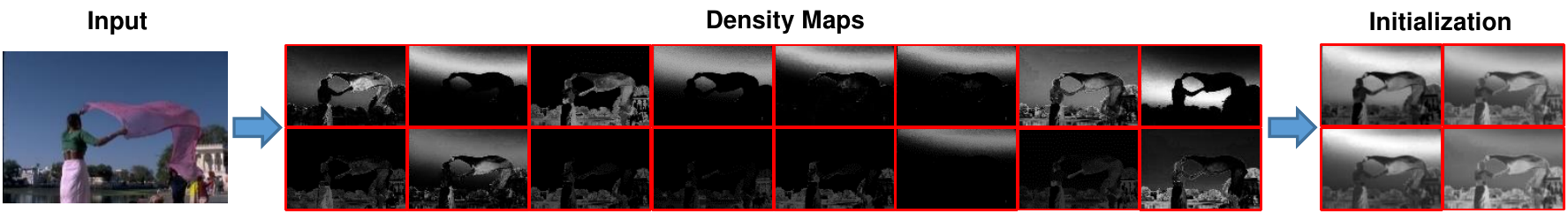}\vspace{-3mm}
\caption{
Initialization. The 16 probability density maps (\textbf{Middle}) are defined with Gaussian models of 16 pixel clusters formed via spectral clustering. Each channel in the initialization (\textbf{Right}) represents the mixed density of 8 Gaussian models. The $i$-th channel (ordered from left to right and top to bottom) mixes 8 density maps of $\{(2^{5-i}j+k)\% 16\,|\,j=1,\cdots,2^{i-1};\,k=1,\cdots,2^{4-i}\}$.
}\vspace{-1mm}
\label{fig:initialization}
\end{figure*}

\subsection{Segmentation by Clustering}
\noindent The first segmentation framework we have considered is based on spectral clustering~\cite{malik2001contour}. Spectral clustering first computes a set of generalized eigenvectors from the locally connected graph. The elements in these eigenvectors corresponding to the same pixel form a $d$-dimensional feature vector at that pixel. It then runs standard clustering, such as K-means, on the pixelwise feature vectors.
A variant of spectral clustering, named weighted spectral clustering, reweighs the $i$-th feature coordinates by $1/\sqrt{\lambda_i}$, where $\lambda_i$ is the eigenvalue corresponding to the $i$-th eigenvector.


Our revised clustering-based segmentation simply replaces the eigenvectors in spectral clustering with the channels in our embedding. That is, each row of the matrix $\mathbf{Y}$ becomes the $d$-dimensional feature vector of the corresponding pixel. Since our embedding is not a spectral one, instead of eigenvalues, we make use of the channel-wise residual cost in our channel weighting scheme defined in (\ref{eq:weighting}). Afterwards, we still perform K-means clustering on the original or weighted channels.

Since our embedding is piecewise flat, pixels of the same region are tightly distributed in the feature space. In contrast, existing embedding techniques, such as Laplacian Eigenmaps, do not share this property. A Laplacian eigenmap is obtained by minimizing the smooth $L_{2,2}$-regularized objective function in (\ref{eq:LEobjs}). The resulting eigenmaps are piecewise smooth but not piecewise flat. Therefore pixels from the same region may still have reasonably large distances among them in the feature space, and may not be grouped together into the same cluster. This is the reason why large or elongated regions are often broken up into pieces in segmentation results based on spectral clustering.

\subsection{Contour-Driven Hierarchical Segmentation}
\noindent One strategy avoiding the drawback of spectral clustering when applied in OWT-UCM~\cite{arbelaez2011contour}, derives global contour information from the eigenmaps computed from the affinity matrix, and then forms segments with respect to contours using hierarchical agglomerative clustering. The hierarchical tree structure built during clustering makes it flexible to choose a segmentation granularity.
To make the algorithm robust, this method also combines local edges with global contours when constructing the contour probability map.
Our contour-driven segmentation follows the same pipeline, except that global contours are extracted from the channels of our embedding instead of the eigenmaps.

Again, since our embedding is piecewise flat, channels in our embedding have almost zero gradients everywhere except at region contours where gradients have a large magnitude. Contour maps extracted from our embedding are both clear and clean without many spurious edges.

\subsection{PFE Initialization for Image Segmentation}\label{sec:ini}
\noindent Initialization plays an important role for the iterative algorithm to converge to a local minimum. In~\cite{yu2015piecewise}, we proposed an initialization scheme based on Gaussian Mixture Models: we fit a Gaussian model to each of the $2^d$ clusters, and use the Gaussians to generate $2^{d}$ density maps. the density maps are then encoded into $d$ images as the initial channels of piecewise flat embedding. Suppose the Gaussians have been ordered into a linear sequence. Each initial channel is formed by summing up $2^{d - 1}$ density maps and subtracting the mean from every pixel. For example, the first channel records the mixed density of the first half of the Gaussians; then we divide the sequence in the middle into two equal subsequences, and the second channel records the mixed density of the first half of the Gaussians in both subsequences. Once the initial $\mathbf{Y}$ has been set, $\mathbf{P}$ is initialized with an orthogonalized version of $\mathbf{D}^{1/2}\mathbf{Y}$ using (\ref{eq:ortho}). An example of initialization is shown in Figure \ref{fig:initialization}.

However, it is neither efficient nor robust as GMM was implemented using expectation maximization iterations and initialized randomly. Here we devise two more effective schemes: hierarchical clustering, and weighted spectral clustering~\cite{Belongie1998} (WSC). 1) For hierarchical clustering, 
we use finest level superpixel from~\cite{felzenszwalb2004efficient}, representing each superpixel with its average RGB values, and agglomeratively merging them until $2^d$ clusters are left. 2) For WSC, we employ RGB values to compute the affinity matrix and cluster pixels into $2^d$ groups. Our final model adopts this WSC-based scheme, and we will evaluate and compare the performances of the three schemes in later section. Another two schemes are also tested to show the importance of initialization: (1) random values; (2) simple color combination, $\mathrm{R}\circ \mathrm{G}$, $\mathrm{R}\circ \mathrm{B}$, $\mathrm{G}\circ \mathrm{B}$, $\mathrm{R}\circ \mathrm{G}\circ \mathrm{B}$ ($\circ$ is element-wise multiplication).

%

\begin{table}[t]
\begin{minipage}{\columnwidth}
\centering
\fontsize{7.5pt}{10pt}
\selectfont
\setlength\tabcolsep{3.5pt}
\caption{Comparison of swPFE$_{0.8}$, Normalized Cut (NCut), spectral clustering (SC) and its weighted version (WSC) on BSDS500 using affinity in~\cite{shi2000normalized}.}\label{tab:rgbaffinity}\vspace{-3mm}
\begin{tabular}{c||cc||cc||cc||cc}
\multirow{2}{*}{method} &\multicolumn{2}{c||}{$F_{op}$} &\multicolumn{2}{c||}{Covering} & \multicolumn{2}{c||}{PRI} & \multicolumn{2}{c}{VI} \\ \cline{2-9}
& fixed & dyn & fixed & dyn & fixed & dyn & fixed & dyn \\ \hline
NCut       &0.126 &0.194 & 0.333 & 0.405 & 0.748 & 0.757 & 2.772 & 2.388 \\ 
SC         &0.146 &0.210 & 0.363 & 0.445 & 0.751 & \textcolor{black}{\bf 0.768} & 2.680 & 2.243 \\ 
WSC\cite{Belongie1998}        &0.160 &0.206 & 0.369 & 0.444 & 0.750 & 0.767 & 2.628 & 2.207 \\ \hline
swPFE$_{0.8}$      &\textcolor{black}{\bf 0.247} &\textcolor{black}{\bf 0.256} &\textcolor{black}{\bf 0.500} &\textcolor{black}{\bf 0.518} &\textcolor{black}{\bf 0.753} & 0.758 &\textcolor{black}{\bf 1.972} &\textcolor{black}{\bf 1.863}\\ 
\end{tabular} \vspace{2mm}
\end{minipage}\\%
\begin{minipage}{\columnwidth}
\centering
\fontsize{7.5pt}{10pt}
\selectfont
\setlength\tabcolsep{1.5pt}
\caption{Comparison of swPFE$_{0.8}$, spectral clustering (SC) and its weighted version (WSC) on BSDS500 using affinity computed with DB as in~\cite{arbelaez2011contour}.
}\label{tab:hedaffinity}\vspace{-3mm}
\begin{tabular}{c||cc||cc||cc||cc||cc}
\multirow{2}{*}{method} & \multicolumn{2}{c||}{$F_{b}$} & \multicolumn{2}{c||}{$F_{op}$} &\multicolumn{2}{c||}{Covering} & \multicolumn{2}{c||}{PRI} & \multicolumn{2}{c}{VI} \\ \cline{2-11}
& fixed & dyn & fixed & dyn & fixed & dyn & fixed & dyn & fixed & dyn \\ \hline
SC            &0.599    &0.641	&0.178	&0.257	&0.378	&0.490	&0.764	&0.788	&2.517	&1.988   \\ 
WSC\cite{Belongie1998}           &0.599	&0.643	&0.185	&0.259	&0.378	&0.479	&0.765	&0.784	&2.496	&1.980   \\ \hline 
swPFE$_{0.8}$ &\textcolor{black}{\bf 0.704}  &\textcolor{black}{\bf 0.716} &\textcolor{black}{\bf 0.426} &\textcolor{black}{\bf 0.428} &\textcolor{black}{\bf 0.643} &\textcolor{black}{\bf 0.653} &\textcolor{black}{\bf 0.847} &\textcolor{black}{\bf 0.848} &\textcolor{black}{\bf 1.388} &\textcolor{black}{\bf 1.343} \\ 
\end{tabular}
\end{minipage}\vspace{-5mm}
\end{table}

\section{Experiments}
\begin{figure*}[t]
\begin{tabular}{p{0.122\linewidth}<{\centering}p{0.122\linewidth}<{\centering}p{0.122\linewidth}<{\centering}p{0.122\linewidth}<{\centering}p{0.122\linewidth}<{\centering}p{0.122\linewidth}<{\centering}p{0.122\linewidth}<{\centering}}
Input & NCut & SC & WSC & gwPFE$_{0.8}$ & hwPFE$_{0.8}$ & swPFE$_{0.8}$\\
\end{tabular}
\centering
\includegraphics[width=\linewidth]{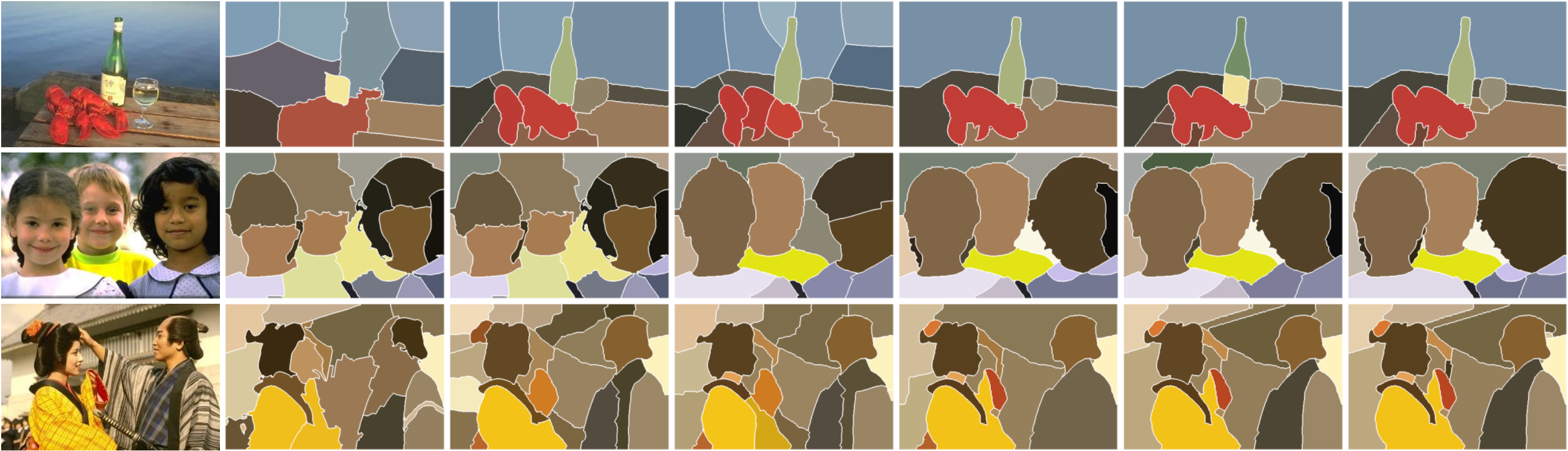} 
\vspace{-5mm}
\caption{Comparison of clustering based segmentation methods. Normalized Cut is based on the original local boundary cues; the other methods adopt local boundary cues from DB~\cite{kokkinos2015pushing}. All the previous methods tend to produce segments that break a semantically coherent region. The results of three different initialization options of our method are shown.}\vspace{0mm}
\label{fig:ClusSeg}
\end{figure*}



\noindent We have conducted experiments within the two aforementioned frameworks: (a) segmentation by clustering and (b) contour-driven hierarchical segmentation, and compared revised segmentation methods incorporating our embedding against the original and other existing methods.

We first incorporate our piecewise flat embedding into the clustering-based segmentation approach, and compare it against three existing spectral methods. Two types of affinity matrices are used: the affinity used in the original Normalized Cut~\cite{shi2000normalized} shown in Table~\ref{tab:rgbaffinity}, and that using local boundary probability~\cite{arbelaez2011contour} shown in Table~\ref{tab:hedaffinity}. 

Clustering-based segmentation using K-means need to specify cluster numbers. We used the following two schemes during performance testing: the \emph{fixed} scheme and the \emph{dynamic} scheme. The \emph{fixed} scheme uses the number of segments provided by the groundtruth segmentation. If there are multiple groundtruth segmentations, we run the algorithm with the number of segments for every groundtruth segmentation, and take the average. The \emph{dynamic} scheme sets the number of segments to odd numbers between 5 and 25 (empirically), and chooses the one with the best performance.
For methods that require multiple NCut eigenvectors, we run each method using 4, 8, and 16 eigenvectors respectively, and take the one with the highest performance.

We have also incorporated piecewise flat embedding into contour-driven hierarchical segmentation following the pipeline in gPb-owt-ucm~\cite{arbelaez2011contour} and MCG~\cite{pont2017multiscale}. In our revised algorithms, we replace the eigenmaps with our piecewise flat embedding. We have also conducted segmentation experiments by additionally replacing the local contours in gPb-owt-ucm~\cite{arbelaez2011contour} with those computed from Deep Boundaries (DB)~\cite{kokkinos2015pushing} and multi-scale boundaries of COB~\cite{Man+17}. We compare the segmentation performance of our revised algorithms with the original or LE-based algorithms and a number of existing algorithms. The comparison results are shown in Tables~\ref{tab:BSDS500benchmark}, \ref{tab:SBDbenchmark}, \ref{tab:MSRCbenchmark} and \ref{tab:PASCALbenchmark}.

\vspace{1.5mm}
\noindent \textbf{Metrics}
The results are evaluated using four standard criteria: $F$ measure for boundaries ($F_{b}$), Segmentation Covering (Covering), Probabilistic Rand Index (PRI), and Variation of Information (VI) as summarized in~\cite{arbelaez2011contour}, plus another widely used metric, $F$ measure for objects and parts ($F_{op}$) proposed in\cite{ponttuset2016supervised}. 
We use threshold values $\{\frac{x}{101}|x=0,1,...,100 \}$ for all UCM based methods.

\vspace{1.5mm}
\noindent \textbf{Datasets}
Experiments and comparison are conducted on the following 4 benchmark datasets:
\begin{itemize}
 \item \textbf{BSDS500}~\cite{arbelaez2011contour}: 200 training, 100 validation and 200 testing images. 5-10 groundtruth segmentations are annotated by different subjects for each image. Image size is 321$\times$481 or 481$\times$321.
 \item \textbf{SBD}: Stanford Background Dataset~\cite{gould2009decomposing}, 715 images of outdoor scenes. 'layers' is used as groundtruth.
 \item \textbf{MSRC}~\cite{shotton2006textonboost}: 591 images and 23 object classes. Precise groundtruth segmentations are provided in \cite{malisiewicz2007improving}.
 \item \textbf{PASCAL Context}~\cite{mottaghi2014role}: 10,103 images and 459 object classes. It is splited into 4,998 training images, 2,607 validation images and 2,498 testing images in~\cite{Man+17}.
\end{itemize}
Object regions are used as groundtruth segmentation without semantic meaning for MSRC and PASCAL Context.

\vspace{1.5mm}
\noindent \textbf{Implementation}
When computing our piecewise flat embedding in the clustering-based segmentation, we set the parameters $\lambda$ and $r$ to 40000 and 600 respectively in stage I, and $r$ is reduced to 10 in stage II. $\alpha = 0.1$, $\varepsilon = 10^{-5}$, and the neighborhood radius is set to 3 in PFE computed using the $L_{1,1}$-regularized and $L_{1,p}$-regularized objective functions, respectively. When computing our piecewise flat embedding in the contour-driven hierarchical segmentation, we set the parameters $\lambda$ and $r$ to 4000 and 600 respectively in stage I, and reduce $r$ to 10 in stage II. $\alpha = 50$, $\varepsilon =  10^{-2}$, and the neighborhood radius is set to 5.

\vspace{1.5mm}
\noindent \textbf{Notations}
For clarity, we define the following expression to differentiate among variants of piecewise flat embedding: [init][weighting?]PFE$_{[\text{norm}]}^{[\text{size}]}$, where [init] is either 'r', 'c', 'g', 'h', or 's', denoting initialization based on random numbers, color combination, GMM, hierarchical clustering or WSC (the new scheme); [weighting?] is either 'w' or 'o', denoting whether residual-based channel weighting is turned on or not; [size] is the width of neighborhood window(default value is 11); and [norm] is either '1' or 'p', '1' for the $L_{1,1}$-regularized objective and 'p' for the $L_{1,p}$-regularized $(p\neq1)$ objective. For example, goPFE$_1$, the same version as in~\cite{yu2015piecewise}, meaning PFE is computed using the $L_{1,1}$-regularized objective in (\ref{eq:PFEobjL11}) and GMM based initialization without channel weighting. And swPFE$_p$ means PFE is computed using the $L_{1,p}$-regularized objective in (\ref{eq:PFEobjL1p}), residual-based channel weighting and the WSC initialization.

Figure~\ref{fig:ClusSeg} shows segmentation results by our method and comparison with algorithms used in our evaluation.

\begin{figure*}[t]
  \begin{minipage}{\linewidth}
  \centering
    \includegraphics[width=0.246\linewidth]{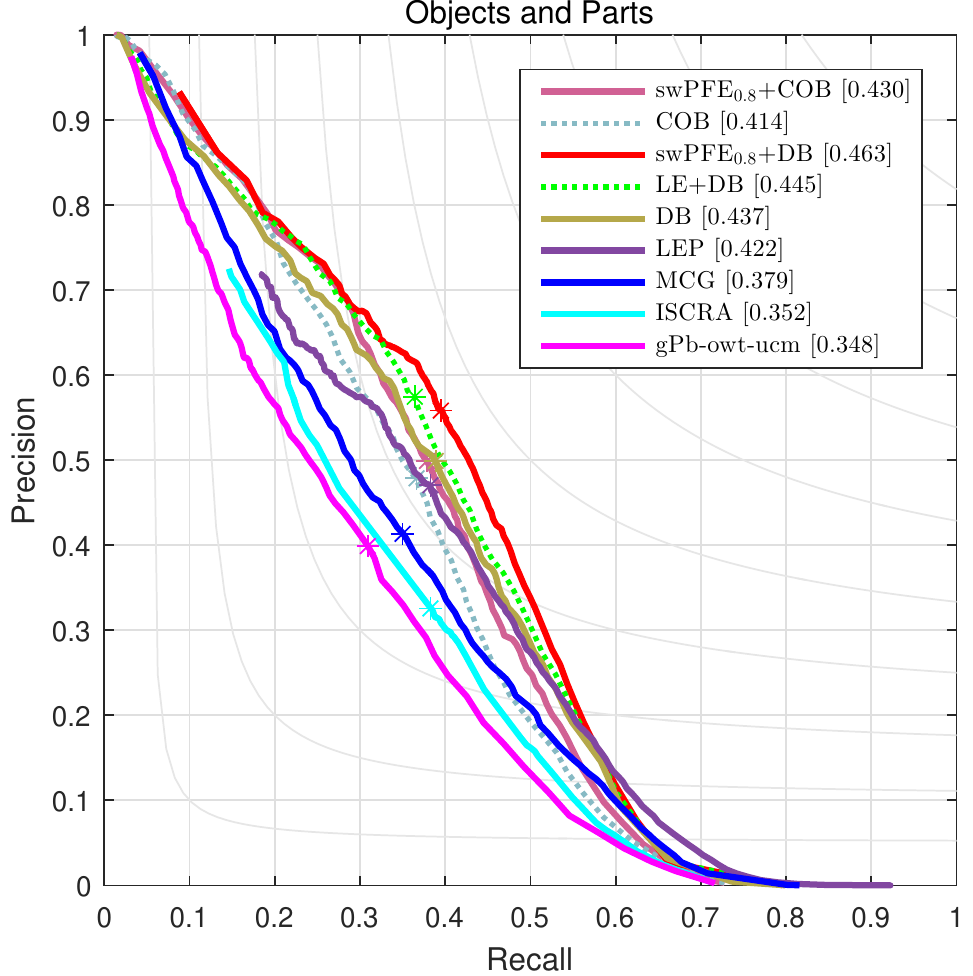}
    \includegraphics[width=0.246\linewidth]{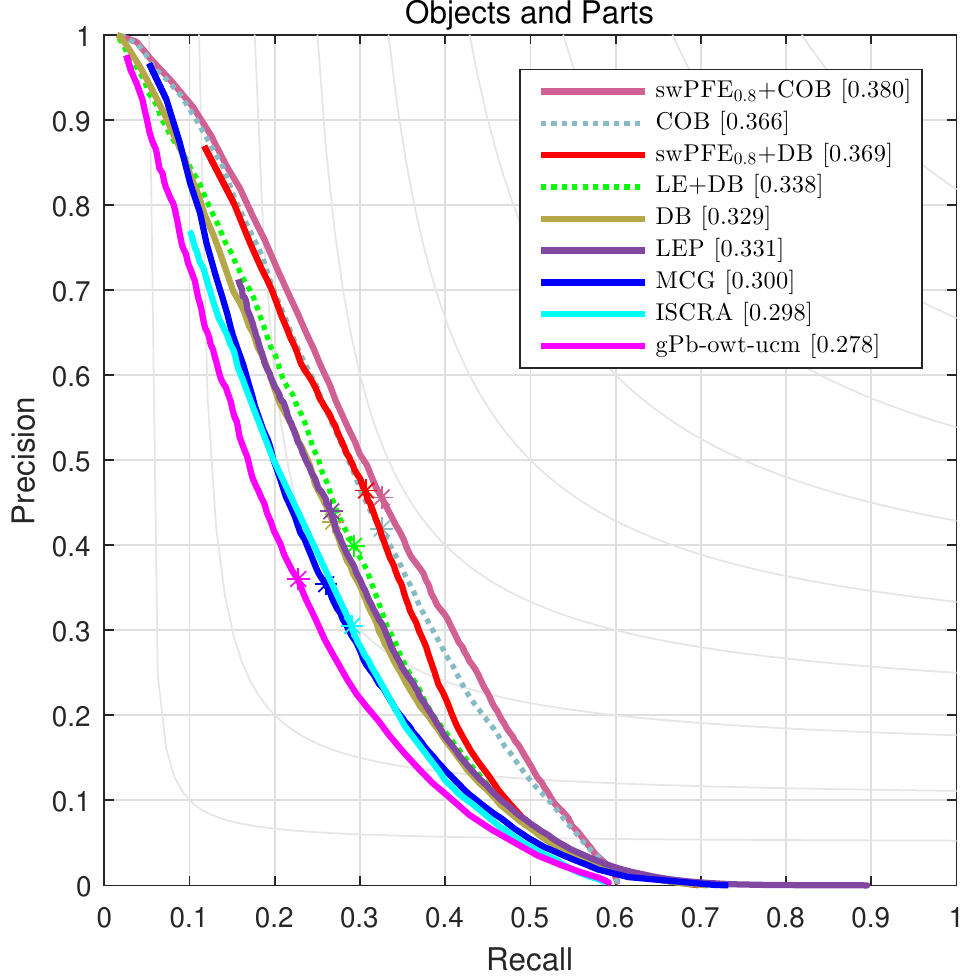}
    \includegraphics[width=0.246\linewidth]{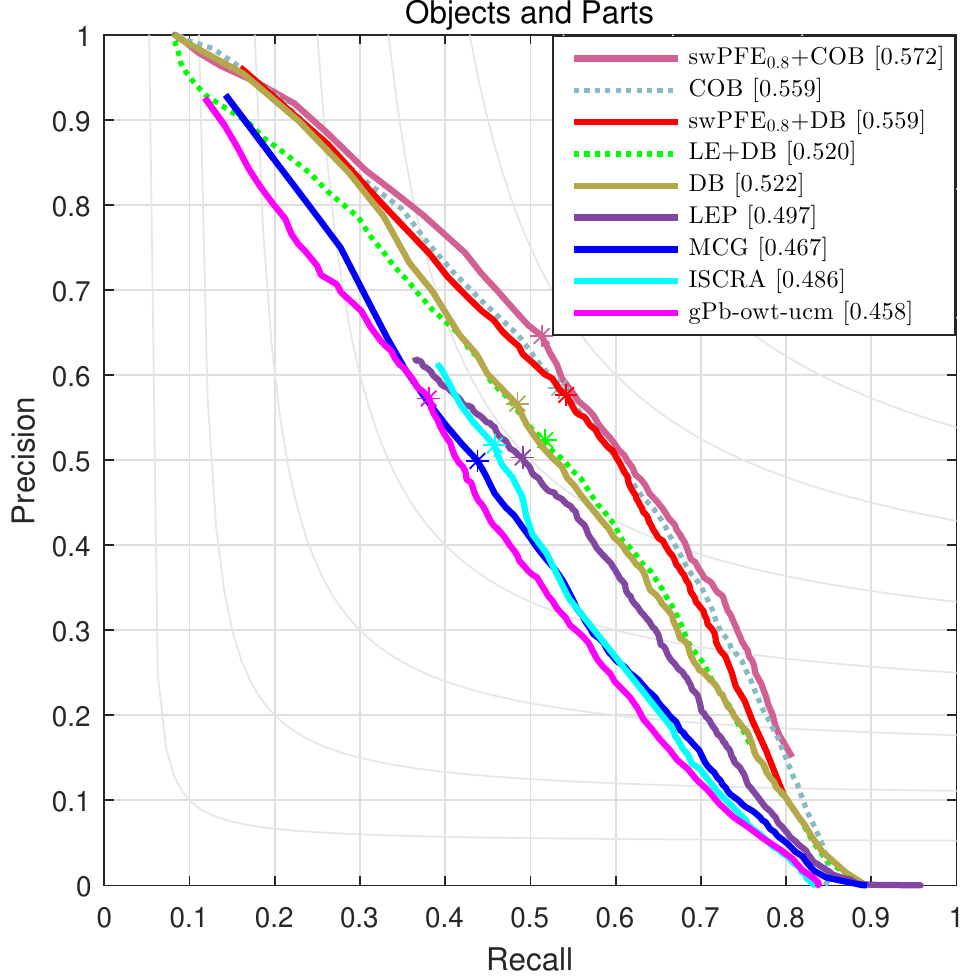}
    \includegraphics[width=0.246\linewidth]{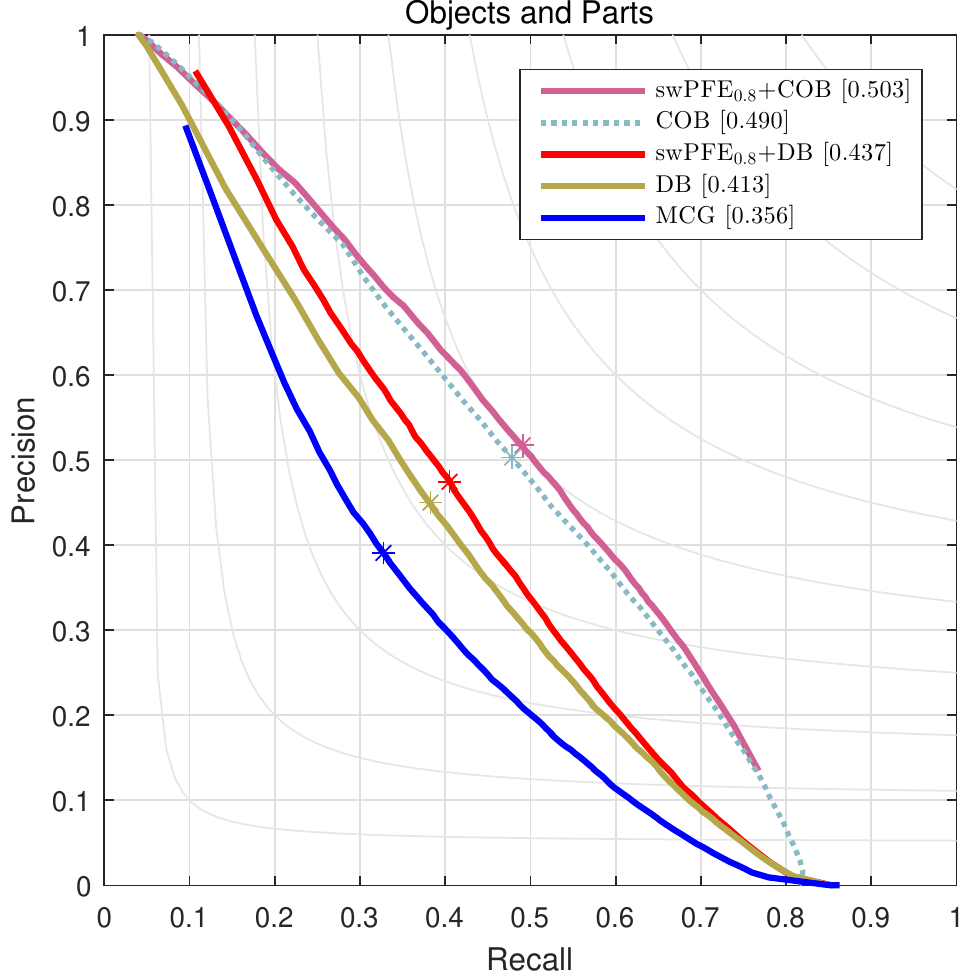}
  \end{minipage}\vspace{0mm}
  \caption{
  Precision-recall curves for the objects and parts measure $F_{op}$. From left to right is result on the BSDS500, SBD, MSRC and PASCAL Context dataset, respectively. Methods using swPFE$_{0.8}$ consistently outperforms original or other existing methods. 
  }\label{fig:fopcomp}\vspace{-3mm}
\end{figure*}

\begin{table}[t]
    \caption{Hierarchical segmentation performance on BSDS500. Segmentation methods integrating swPFE$_{0.8}$
     perform favorably over their original or LE-based counterparts (all implemented in OWT-UCM) and other methods. Here the integration of powerful local boundary cues, i.e. DB, is critical for our method to achieve state-of-the-art performance.
    }\label{tab:BSDS500benchmark}\vspace{-1mm}
    \centering
    \fontsize{7pt}{10pt}
    \selectfont
    \setlength\tabcolsep{1.4pt}
    \begin{tabular}{ r||cc||cc||cc||cc||cc }
  \multicolumn{1}{c||}{\multirow{2}{*}{method}} & \multicolumn{2}{ c|| }{$F_{b}$}  & \multicolumn{2}{ c|| }{$F_{op}$} & \multicolumn{2}{ c|| }{Covering} & \multicolumn{2}{ c|| }{PRI} & \multicolumn{2}{ c }{VI} \\ \cline{2-11}
 &ODS & OIS &ODS & OIS & ODS & OIS & ODS & OIS & ODS & OIS \\ \hline
    MS-NCut~\cite{cour2005spectral}       &0.64      &0.68    &0.213     &0.270   &0.45 & 0.53 & 0.78 & 0.80 & 2.23 & 1.89 \\ 
    Mean Shift~\cite{comaniciu2002mean}   &0.64  &0.68  &0.229  &0.292  & 0.54 & 0.58 & 0.79 & 0.81 & 1.85 & 1.64 \\ 
    Hoiem~\cite{hoiem2011recovering}      &-&-&-&-& 0.56 & 0.60 & 0.80 & 0.77 & 1.78 & 1.66 \\ 
    ISCRA~\cite{ren2013image}             &0.717 &0.752 &0.352 &0.418 &0.592 &0.660  & 0.821 & 0.857 & 1.605 & 1.415 \\ 
     1-SC~\cite{hein2010inverse}
    &0.734 &0.775 &0.364 &0.455 &0.569 &0.655 &0.811 &0.854 &1.613 &1.323 \\
    LEP~\cite{zhaosegmenting}
    &0.754 &0.789 &0.422 &0.466 &0.629 &0.677 &0.836 &0.864 &1.468 &1.325\\ 

    \hline
　  gPb-owt-ucm~\cite{arbelaez2011contour}&0.729 &0.755 &0.348 &0.385 &0.587 &0.646  & 0.828 & 0.855 & 1.690 & 1.476 \\ 
　  swPFE$_{0.8}$+mPb                    &0.723 &0.752  &0.364 &0.411 &0.612 &0.669 &0.830 &0.866 &1.565 &1.386 \\ \hline　
    MCG~\cite{pont2017multiscale}     &0.744 &0.777 &0.379 &0.433 &0.613 &0.663  & 0.832 & 0.861 & 1.568 & 1.390 \\ 
    swPFE$_{0.8}$+MCG                         &0.747 &0.779 &0.397 &0.448 &  0.622 & 0.679 & 0.838 & 0.870 & 1.557 & 1.365 \\ \hline　
    DB~\cite{kokkinos2015pushing}                                  &\textcolor{black}{\bf0.809} & 0.832 &0.437    &0.491 &0.663 &0.714 &0.857  &0.886  &1.387 &1.211 \\ 
    WLS~\cite{farbman2008edge}+DB
    &0.808 &0.828 &0.440 &0.490 &0.663 &0.711 &0.857 &0.882 &1.379 &1.214 \\
    SDF~\cite{ham2015robust}+DB
    &{\bf0.809} &0.828 &0.441 &0.495 &0.669 &0.714 &0.859 &0.885 &1.361 &1.207 \\
    LE+DB                                  &\textcolor{black}{\bf0.809} &\textcolor{black}{\bf 0.833} &0.445	&0.500 &0.665 &0.716 &0.858	 &0.886	 &1.365	&1.197 \\ 
    swPFE$_{0.8}$+DB                  &\textcolor{black}{\bf0.809} &0.832 &\textcolor{black}{\bf 0.463} &\textcolor{black}{\bf 0.514} &\textcolor{black}{\bf 0.680} &\textcolor{black}{\bf 0.722} &\textcolor{black}{\bf 0.860} &\textcolor{black}{\bf 0.887} & \textcolor{black}{\bf 1.293} &\textcolor{black}{\bf 1.145} \\ \hline
    COB~\cite{Man+17}
    &0.782 &0.808 &0.414 &0.464 &0.664 &0.712 &0.854 &0.886 &1.380 &1.222\\
    LE+COB
    &0.782 &0.809 &0.413 &0.462 &0.656 &0.706 &0.854 &0.882 &1.384 &1.245\\
    swPFE$_{0.8}$+COB
    &0.787 &0.810 &0.430 &0.485 &0.670 &0.718 &0.858 &0.887 &1.337 &1.182\\
\end{tabular}\vspace{-2mm}
\end{table}

\begin{table}[t]
    \caption{Hierarchical segmentation performance on SBD dataset. Our method performs favorably against its LE and previous methods. \vspace{-1mm}
    }\label{tab:SBDbenchmark}
    \centering
    \fontsize{7pt}{10pt}
    \selectfont
    \setlength\tabcolsep{1.4pt}
    \begin{tabular}{ r||cc||cc||cc||cc||cc }
  \multicolumn{1}{c||}{\multirow{2}{*}{method}} & \multicolumn{2}{ c|| }{$F_{b}$}  & \multicolumn{2}{ c|| }{$F_{op}$} & \multicolumn{2}{ c|| }{Covering} & \multicolumn{2}{ c|| }{PRI} & \multicolumn{2}{ c }{VI} \\ \cline{2-11}
  &ODS & OIS &ODS & OIS & ODS & OIS & ODS & OIS & ODS & OIS \\ \hline
 gPb-owt-ucm~\cite{arbelaez2011contour}  &0.534 &0.574 &0.279 &0.321 &0.583 &0.643  &0.862 &0.893 &1.877 &1.617 \\ 
 ISCRA~\cite{ren2013image}               &0.539 &0.582 &0.298 &0.340 &0.616 &0.676  &0.871 &0.901 &1.725 &1.493 \\ 
 MCG~\cite{pont2017multiscale}       &0.543 &0.585 &0.300 &0.345 &0.598 &0.663  &0.870 &0.897 &1.756 &1.501 \\ 
 LEP\cite{zhaosegmenting}
 &0.550 &0.597 &0.331 &0.374 &0.620 &0.689 &0.874 &0.903 &1.629 &1.399\\
 \hline

 DB~\cite{kokkinos2015pushing}
 &0.631   &0.667  &0.329  &0.375 &0.667 &0.724 &0.890 &0.920 &1.532 &1.314 \\
 LE+DB
 &0.633   &0.670  &0.336  &0.379 &0.663 &0.720 &0.889 &0.918 &1.510 &1.302 \\ 
 swPFE$_{0.8}$+DB
 & 0.634	&0.669	&0.369	 &0.407	 &0.681	 &0.734	 &0.892	 &0.922	 &1.403	 &1.229 \\ \hline
 COB\cite{Man+17}
 &\textcolor{black}{\bf0.666} &\textcolor{black}{\bf0.693} &0.366 &0.416 &0.721 &0.766 &0.910 &0.934 &1.324 &1.164\\
 LE+COB
 &0.663 &0.689 &0.371 &0.417 &0.722 &0.764 &0.911 &0.933 &1.308 &1.168 \\
 swPFE$_{0.8}$+COB
 &0.663 &0.690 &\textcolor{black}{\bf0.380} &\textcolor{black}{\bf0.432} &\textcolor{black}{\bf0.728} &\textcolor{black}{\bf0.772} &\textcolor{black}{\bf0.911} &\textcolor{black}{\bf0.935} &\textcolor{black}{\bf1.292} &\textcolor{black}{\bf1.131}\\
\end{tabular}\vspace{-3mm}
\end{table}

\begin{table}[t]
    \caption{Hierarchical segmentation performance on MSRC dataset. Our method performs favorably against its LE and previous methods. \vspace{-1mm}
    }\label{tab:MSRCbenchmark}
    \centering
    \fontsize{7pt}{10pt}
    \selectfont
    \setlength\tabcolsep{1.4pt}
    \begin{tabular}{ r||cc||cc||cc||cc||cc }
 \multicolumn{1}{c||}{\multirow{2}{*}{method}} & \multicolumn{2}{ c|| }{$F_{b}$}  & \multicolumn{2}{ c|| }{$F_{op}$} & \multicolumn{2}{ c|| }{Covering} & \multicolumn{2}{ c|| }{PRI} & \multicolumn{2}{ c }{VI} \\ \cline{2-11}
 &ODS & OIS &ODS & OIS & ODS & OIS & ODS & OIS & ODS & OIS \\ \hline
 gPb-owt-ucm~\cite{arbelaez2011contour} &0.468 &0.524 &0.458 &0.525 &0.651 &0.743  &0.779 &0.845 &1.274 &0.980 \\ 
 ISCRA~\cite{ren2013image}              &0.463 &0.530 &0.486 &0.524 &0.667 &0.746  &0.767 &0.845 &1.166 &1.024 \\ 
 MCG~\cite{pont2017multiscale}      &0.479 &0.537 &0.467 &0.536 &0.662 &0.745  &0.784 &0.848 &1.205 &0.961 \\ 
 LEP\cite{zhaosegmenting}
 &0.499 &0.553 &0.497 &0.538 &0.677 &0.741 &0.792 &0.843 &1.190 &1.051\\
 \hline
 DB~\cite{kokkinos2015pushing}                                  &0.570  &0.623  &0.522  &0.608  &0.733  &0.808  &0.825  &0.888  &1.061  &0.806 \\
 LE+DB                                  &0.586	&0.630	&0.536	&0.619	&0.734	&0.807	&0.827	&0.887	&1.012	&0.793 \\ 
 swPFE$_{0.8}$+DB
 &0.587	&0.632	 &0.559	 &0.634	 &0.748	 &0.816	 &0.835	 &0.891	 &0.970 &0.761 \\ \hline
 COB\cite{Man+17}
 &0.597 &0.634 &0.559 &0.635 &0.764 &0.823 &0.848 &0.895 &0.966 &0.760\\
 LE+COB
 &0.591 &0.630 &0.559 &0.621 &0.757 &0.814 &0.845 &0.888 &0.980 &0.783\\
 swPFE$_{0.8}$+COB
 &\textcolor{black}{\bf0.598} &\textcolor{black}{\bf0.636} &\textcolor{black}{\bf0.572} &\textcolor{black}{\bf0.642} &\textcolor{black}{\bf0.767} &\textcolor{black}{\bf0.824} &\textcolor{black}{\bf0.850} &\textcolor{black}{\bf0.893} &\textcolor{black}{\bf0.953} &\textcolor{black}{\bf0.751}  \\
\end{tabular}\vspace{-2mm}
\end{table}

\begin{table}[t]
    \caption{Hierarchical segmentation performance on PASCAL Context dataset. Our method performs favorably against LE and other methods. \vspace{-1mm}
    }\label{tab:PASCALbenchmark}
    \centering
    \fontsize{7pt}{10pt}
    \selectfont
    \setlength\tabcolsep{1.4pt}
    \begin{tabular}{ r||cc||cc||cc||cc||cc }
 \multicolumn{1}{c||}{\multirow{2}{*}{method}} & \multicolumn{2}{ c|| }{$F_{b}$}  & \multicolumn{2}{ c|| }{$F_{op}$} & \multicolumn{2}{ c|| }{Covering} & \multicolumn{2}{ c|| }{PRI} & \multicolumn{2}{ c }{VI} \\ \cline{2-11}
 &ODS & OIS &ODS & OIS & ODS & OIS & ODS & OIS & ODS & OIS \\ \hline
 MCG~\cite{pont2017multiscale}      &0.577 &0.634 &0.356 &0.419 &0.577 &0.668  &0.798 &0.854 &1.680 &1.332 \\ 
 DB~\cite{kokkinos2015pushing}
 &0.680  &0.731  &0.413  &0.489  &0.658  &0.748  &0.840  &0.896  &1.421  &1.112 \\
 LE+DB
 &0.688 &0.736   &0.416  &0.491  &0.663  &0.744  &0.840  &0.893  &1.395  &1.114 \\
 swPFE$_{0.8}$+DB
 &0.691 &0.741 &0.437 &0.511 &0.669 &0.753 &0.841 &0.898 &1.327 &1.060\\ \hline
 COB\cite{zhaosegmenting}
 &0.755 &0.789 &0.490 &0.566 &0.739 &0.803 &0.878 &0.919 &1.150 &0.916\\
 LE+COB
 &0.602 &0.655 &0.350 &0.417 &0.588 &0.675 &0.806 &0.859 &1.693 &1.332\\
 swPFE$_{0.8}$+COB
 &\textcolor{black}{\bf0.757} &\textcolor{black}{\bf0.791} &\textcolor{black}{\bf0.503} &\textcolor{black}{\bf0.578} &\textcolor{black}{\bf0.743} &\textcolor{black}{\bf0.806} &\textcolor{black}{\bf0.880} &\textcolor{black}{\bf0.920} &\textcolor{black}{\bf1.129} &\textcolor{black}{\bf0.894}\\
\end{tabular}\vspace{-3mm}
\end{table}

\subsection{Results and Comparison}
\noindent \textbf{Segmentation Using PFE vs Other Existing Methods}
In the framework of clustering-based segmentation, methods incorporating our piecewsie flat embedding achieve significantly better performance than those adopting Normalized Cuts including spectral clustering and weighted spectral clustering, as shown in Tables \ref{tab:rgbaffinity} and \ref{tab:hedaffinity}. 
Particularly, variant of our embedding, swPFE$_{0.8}$, achieves 0.247 on $F_{op}$ (\emph{fixed} scheme), which is 54.4\% higher than the corresponding method using Laplacian Eigenmaps in Table \ref{tab:rgbaffinity}.
In the context of hierarchical image segmentation, as our embedding is not a standalone segmentation algorithm, we demonstrate that incorporating our embedding makes existing hierarchical algorithms achieve clearly better performance. As shown in Table \ref{tab:BSDS500benchmark}, once PFE is integrated into gPb-owt-ucm, the resulting algorithm, `swPFE$_{0.8}$+mPb', exhibits better performance under all considered metrics except for $F_b$. Its $F_{op}$ (OIS) is increased from 0.385 to 0.411, and VI (ODS) is improved from 1.690 to 1.565. Likewise, `swPFE$_{0.8}$+MCG' delivers better performance than the original MCG under all metrics. After further replacing the local contours in gPb-owt-ucm with those computed using Deep Boundaries~\cite{kokkinos2015pushing}, the resulting `swPFE$_{0.8}$+DB' delivers substantially better performances than its LE counterpart (`LE+DB'), and achieves state-of-the-art performance in BSDS500 under the contour-driven hierarchical segmentation framework. Consistently favorable results are delivered by PFE-based methods in SBD (Table~\ref{tab:SBDbenchmark}), MSRC (Table~\ref{tab:MSRCbenchmark}) and PASCAL Context (Table~\ref{tab:PASCALbenchmark}). Precision-recall curves of various segmentation algorithms for $F_{op}$ measure are shown in Figure~\ref{fig:fopcomp}.

\begin{figure*}[t]
  \centering
    \includegraphics[width=0.3\linewidth]{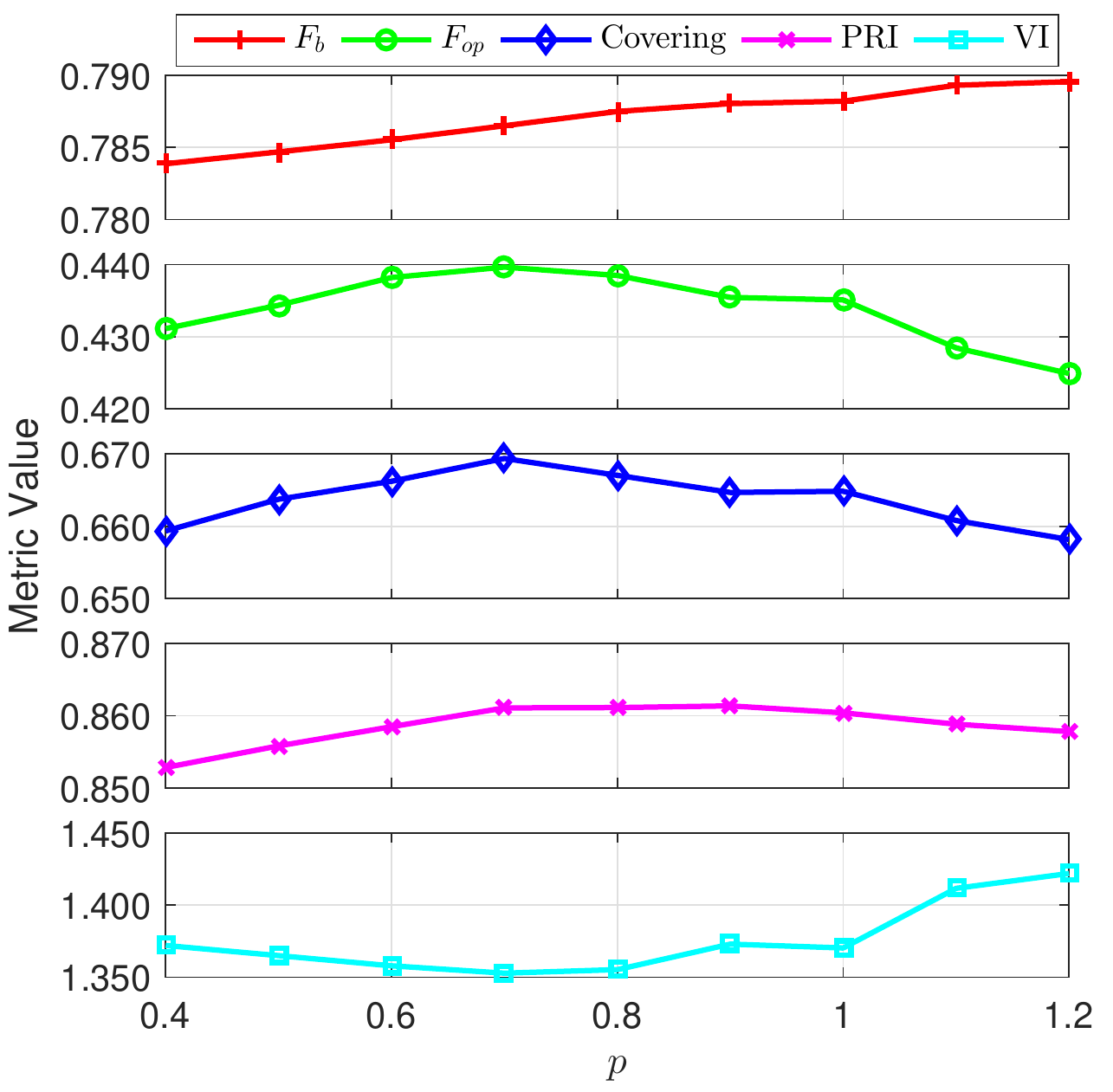} \hfill
    \includegraphics[width=0.3\linewidth]{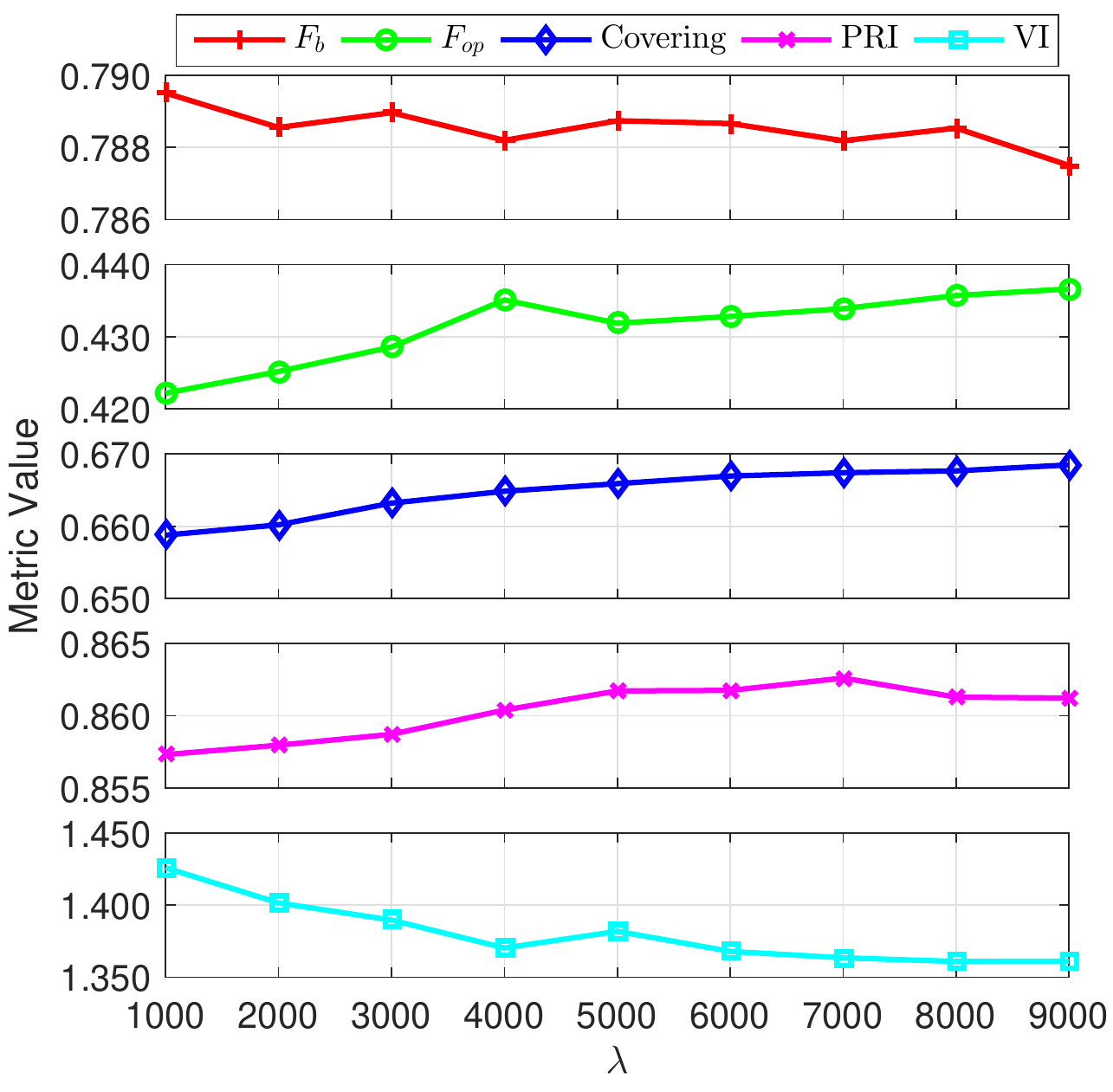} \hfill
    \includegraphics[width=0.3\linewidth]{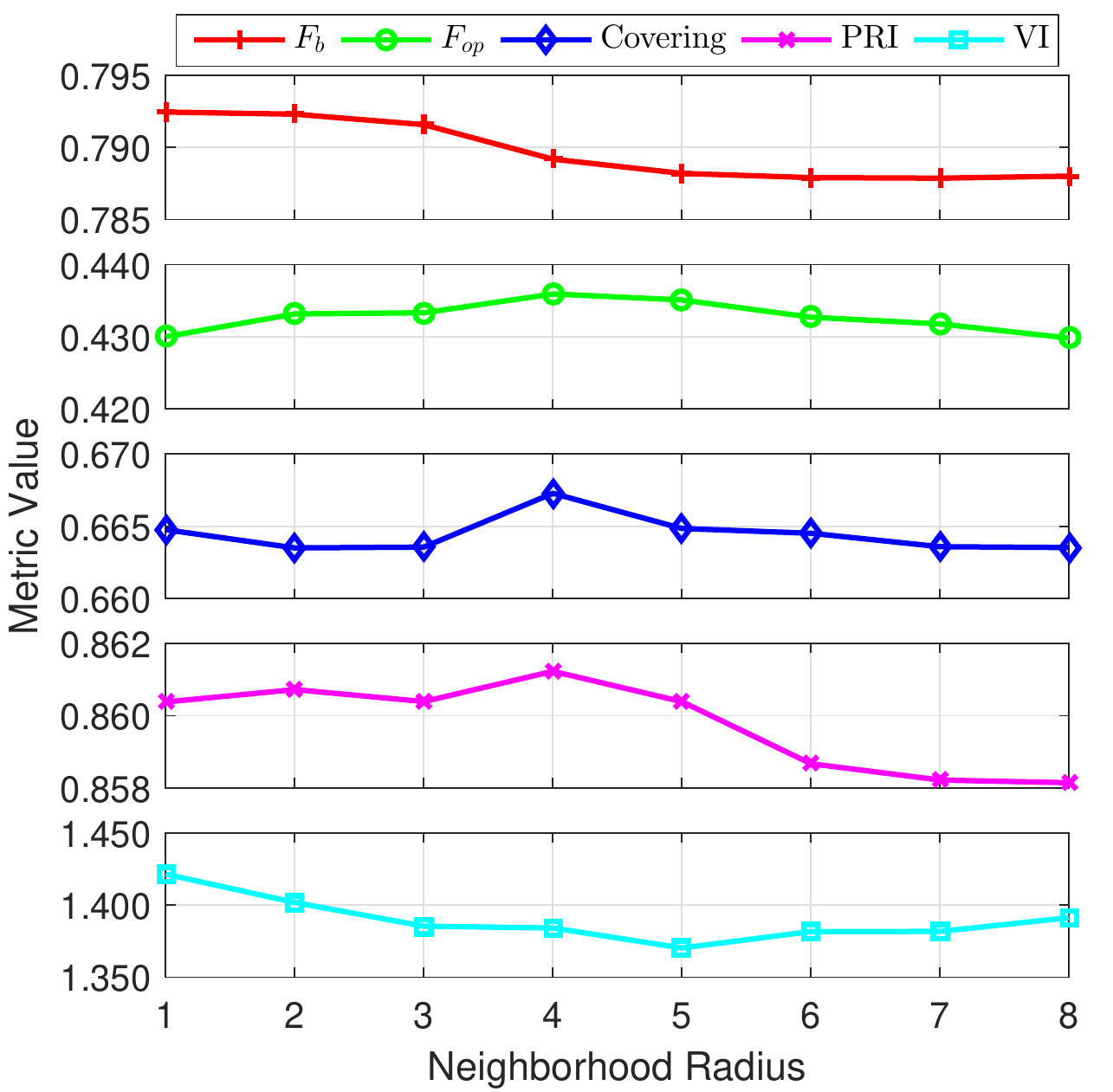}
  \vspace{-3mm}
  \caption{Ablation analysis of parameters ($p$ , $\lambda$, and neighborhood size) on final segmentation performance (ODS values of five measures). In each experiment, all parameters except for the analyzed one are fixed to the default setting. Experiments are conducted on the BSDS500 dataset in the contour-driven hierarchical segmentation framework.
  }\label{fig:parameters}\vspace{-4mm}
\end{figure*}

In Table~\ref{tab:BSDS500benchmark}, performances of MS-NCut~\cite{cour2005spectral} and Mean Shift~\cite{comaniciu2002mean} are taken from~\cite{arbelaez2011contour}. The result of Hoiem~\cite{hoiem2011recovering} is from ~\cite{ren2013image}. Other methods are evaluated through the SEISM toolbox~\cite{ponttuset2016supervised}. The algorithm in \cite{hein2010inverse} (denoted as 1-SC) taking affinity matrices computed from local boundary~\cite{kokkinos2015pushing} is tested as a hierarchical segmentation method. Maximum number of clusters is 60 and normalized criteria is adopted. For COB, we use the precomputed result provided by~\cite{Man+17}. As we mentioned earlier, our objective function bears some resemblance with the energy function (\ref{eq:WNonlocalTV}) which utilizes weighted nonlocal total variation for regularization. We have attempted to make use of it in image segmentation by directly extracting global contours from 3 scales of its smoothed outputs. The affinity measure $W_{ij}$ is computed from DB. The WLS (weighted least squares) method is used to solve it and its performance is reported as 'WLS+DB'. Similarly the filtering method in~\cite{ham2015robust} is also integrated into image segmentation and reported as 'SDF+DB'. However, the results are not comparable to result of our method.

In addition to region-oriented measures, boundary-oriented measure $F_b$ is also widely used to evaluate the performance of image segmentation. Since our embedding focuses on flattening regions and we still use existing local boundary cues in contour-driven segmentation, our piecewise flat embedding tends to substantially improve region-oriented segmentation measures while achieving equivalent or slightly better results under $F_b$.

\vspace{1.5mm}
\noindent \textbf{Initialization Schemes} 
Table~\ref{tab:AblationExper} shows performance perturbation of different initialization schemes, and of weighted vs. original PFE schemes. The initialization schemes are random values, color combination, GMM, hierarchical segmentation, and WSC that we described in Section~\ref{sec:ini}.  In the original PFE scheme, i.e. without the residual-based channel weighting, the three initialization schemes make little differences (the top three rows).  With the weighting scheme, GMM is noticeably worse, and WSC is slightly better than hierarchical-based initialization. For example, $F_{op}$ (ODS) measure of 'gwPFE$_{0.8}$+DB' is $0.44$ while those of the other two schemes are both over $0.46$; the VI (ODS) measure of 'gwPFE$_{0.8}$+DB' is $1.330$, while 'hwPFE$_{0.8}$+DB' is $1.302$ and 'swPFE$_{0.8}$+DB' is $1.293$. Initializations using random values, color combination and GMM derive almost same performances except for different VI measures.  The experiments were conducted on the BSDS500 dataset. Results on the other datasets exhibit similar behavior.

\vspace{1.5mm}
\noindent \textbf{Original PFE vs Weighted PFE}
The residual-based channel weighting makes a clear difference in all three initialization schemes as shown in Table~\ref{tab:AblationExper}. For example, if we compare $F_{op}$ (ODS) measure of 'swPFE$_{0.8}$+DB' against 'soPFE$_{0.8}$+DB', the former is $0.463$ while the latter is $0.438$. On the $F_{op}$ (ODS) measure, the performance gain brought by channel weighting for the GMM, hierarchical and WSC initializations are $1\%$, $6\%$, and $6\%$, respectively. Besides, further experiments show that the weighting scheme has less effect in the clustering-based segmentation framework.

\begin{figure*}[t]
\begin{tabular}{p{0.143\linewidth}<{\centering}p{0.143\linewidth}<{\centering}p{0.143\linewidth}<{\centering}p{0.143\linewidth}<{\centering}p{0.143\linewidth}<{\centering}p{0.143\linewidth}<{\centering}}
Input & gPb-UCM  & MCG & LEP & COB &swPFE$_{0.8}$+COB \\
\end{tabular}
\centerline{\includegraphics[width=\linewidth]{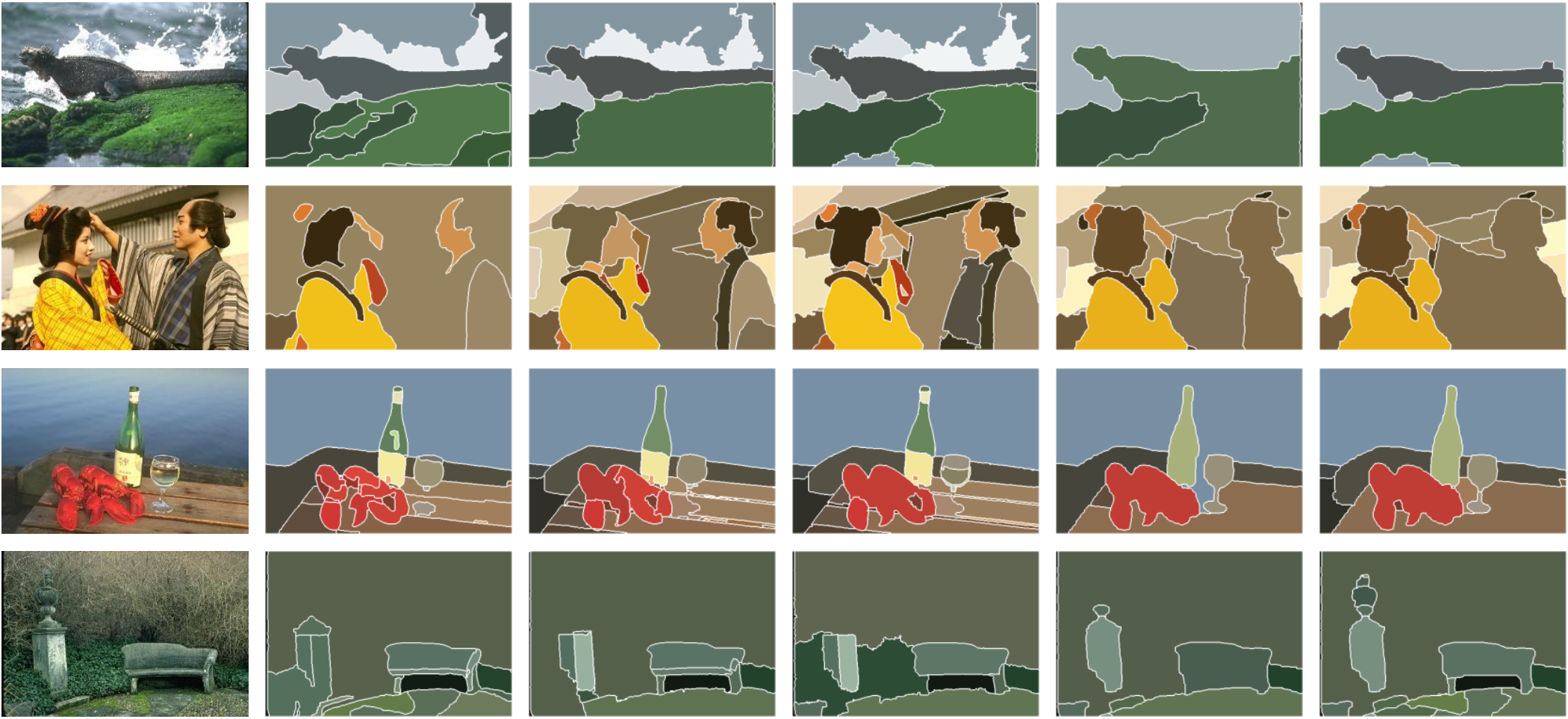}}
\vspace{-3mm}
\caption{Comparison of contour-driven hierarchical segmentation methods on BSDS500. 'swPFE$_{0.8}$+COB' denotes the contour-driven segmentation algorithm that integrates swPFE$_{0.8}$ while utilizing local boundary information from COB.}
\label{fig:HierarSegBSDS500}
\vspace*{\floatsep}
\begin{tabular}{p{0.143\linewidth}<{\centering}p{0.143\linewidth}<{\centering}p{0.143\linewidth}<{\centering}p{0.143\linewidth}<{\centering}p{0.143\linewidth}<{\centering}p{0.143\linewidth}<{\centering}}
Input & Groundtruth &gPt-UCM  & MCG & COB &swPFE$_{0.8}$+COB \\
\end{tabular}
\centerline{\includegraphics[width=\linewidth]{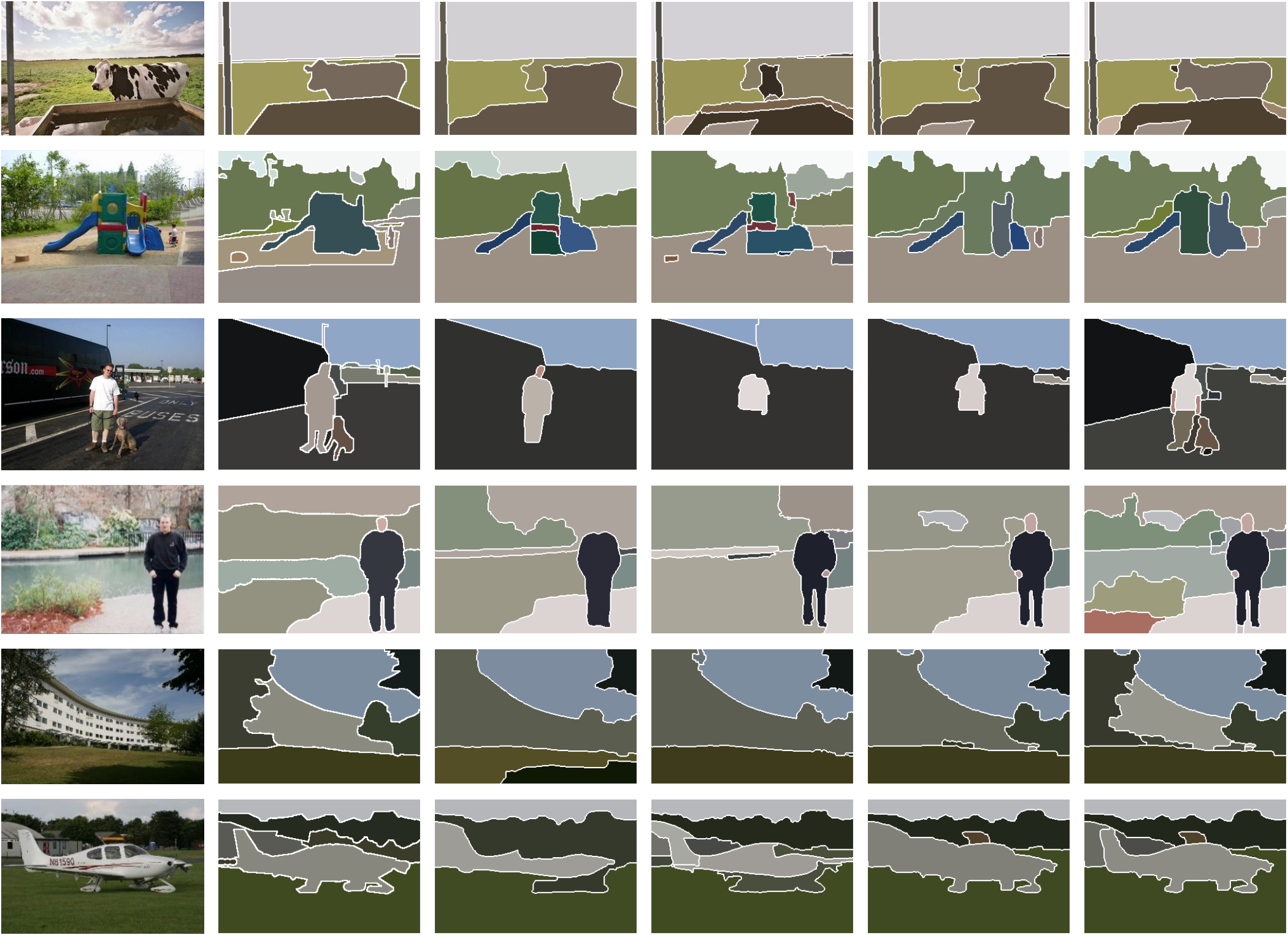}}
\vspace{-3mm}
\caption{Comparison of contour-driven hierarchical segmentation methods on SBD and MSRC. Top 3 images are selected from SBD, and other images are from MSRC.}
\label{fig:HierarSegSBD}
\end{figure*}

A gallery of segmentation results from various algorithms and comparisons can be found in Figures \ref{fig:ClusSeg}, \ref{fig:HierarSegBSDS500} and \ref{fig:HierarSegSBD}.

\begin{table}[t]
\caption{Experiment with various options in the contour-driven hierarchical segmentation task on BSDS500. DB is utilized as local boundary.}\vspace{-1mm} \label{tab:AblationExper}
\centering
\fontsize{7pt}{10pt}
\selectfont
\setlength\tabcolsep{1.8pt}
\begin{tabular}{ r||c|c|c||c|c|c|c||c|c|c }
\hline
Initialization& g   & h   &s   &r           &c           &g           &h           &s           &s           &s \\
Weighting     &     &     &    &$\checkmark$&$\checkmark$&$\checkmark$&$\checkmark$&$\checkmark$&$\checkmark$&$\checkmark$ \\
$p$           &0.8&0.8&0.8&0.8&0.8&0.8&0.8&0.5&0.8&1 \\ \hline
$F_{op}$ ODS  &0.435&0.437&0.438&0.445&0.443&0.440&0.465&0.450&0.463&0.462 \\
Covering ODS  &0.661&0.662&0.663&0.664&0.668&0.665&0.675&0.666&0.680&0.672 \\
PRI ODS       &0.857&0.856&0.856&0.857&0.858&0.857&0.860&0.854&0.860&0.859 \\
VI ODS        &1.393&1.387&1.380&1.364&1.349&1.330&1.302&1.316&1.293&1.315 \\\hline
 \end{tabular}\vspace{-2mm}
\end{table}


\subsection{Parameter Choice}

\noindent In addition to the categorical parameters such as initialization and channel weighting option, let's further investigate the effects of a few important continuous parameters on the final segmentation performance. These parameters include $p$, $\lambda$, and neighborhood size.
We use the contour-driven hierarchical segmentation framework for this evaluation, and take the ODS value of all five performance metrics for this evaluation. The results are plotted in Figure \ref{fig:parameters}.

\noindent \textbf{$\bullet\; p$} controls the sparsity of the embedding. Smaller $p$ improves the sparsity. However a sparser embedding does not always gives rise to better segmentation performance.
In our evaluation, the best result is achieved around $p=0.7$ under all measures except for $F_b$, which improves as $p$ increases.

\noindent \textbf{$\bullet$} As shown in step (a.1) (Appendix~\ref{appendI}), $\lambda$ is the weight of the $L_{1,1}$-regularized energy term. It balances between the sparsity of the embedding and the orthogonality among the embedding channels. The five performance metrics do not change consistently as $\lambda$ varies. It produces the best result with respect to PRI when $\lambda$ is around $7000$. However, $F_{op}$, Covering and VI become better overally as $\lambda$ increases, and $F_b$ remains stable within our testing range with the maximum variation less than $0.003$.


\noindent \textbf{$\bullet$ Neighborhood Size} A larger neighborhood allows direct interaction among pixels further apart, but decreases the sparsity of boundary-crossing connections and brings difficulty in computing accurate pairwise affinity. The best performance on BSDS500 under four measures other than $F_b$ is achieved when the neighborhood radius is around 4, while $F_b$ benefits from smaller neighborhood radius.

\section{Conclusions}
\noindent
We have presented a new multi-dimensional nonlinear embedding called piecewise flat embedding for image segmentation. We adopt an $L_{1,p}$-regularized ($0<p\le 1$) energy term in the objective function to promote sparse solutions. To solve the $L_{1,1}$-regularized objective function, a two-stage numerical algorithm is first developed by nesting two existing solvers. Then we generalize this numerical algorithm through iterative reweighting to solve the $L_{1,p}$-regularized problem. Experiments on BSDS500, MSRC, SBD and PASCAL Context indicate that segmentation methods incorporating our embedding can achieve significantly improved results.



\ifCLASSOPTIONcompsoc
  \section*{Acknowledgments}
\else
  \section*{Acknowledgment}
\fi

This work is supported in part by Hong Kong Research Grants Council under General Research Funds (HKU719313). ZL is supported in part by NSFC Young Researcher Grant (No. 61602406) and by Alibaba-ZJU Joint Institute of Frontier Technologies.


%



\ifCLASSOPTIONcaptionsoff
  \newpage
\fi



\bibliographystyle{IEEEtran}
\bibliography{TPAMIPFEbib}

\clearpage
\appendices
\section{$L_{1,1}$-regularized minimization}\label{appendI}


According to the Split Bregman algorithm~\cite{goldstein2009split}, the $L_{1,1}$-regularized problem in step (a) of equation (\ref{eq:PFEobjL1d1}) can be solved by iterating the following steps:
\begin{itemize}
  \setlength\itemsep{1em}
\item[(a.1)]
  $\mathbf Y^{(k,l+1)}=\arg \underset{\mathbf Y} \min \; \frac{\lambda}{2} \|\mathbf{M} \mathbf Y +\mathbf E^{l} - \mathbf C^{l} \|_{2,2}^2+\\
  	\phantom{Y^{(k,l+1)}=\arg \underset{\mathbf Y} \min \;} \frac{r}{2} ||\mathbf{D}^{1/2} \mathbf Y - \mathbf F^{(k)} ||_{2,2}^2$;
\item[(a.2)] $\mathbf C^{l+1}=\arg \underset{\mathbf C} \min \; \| \mathbf C\|_{1,1}+\frac{\lambda}{2} \| \mathbf C - \mathbf M \mathbf Y^{(k,l+1)} - \mathbf E^l\|_{2,2}^{2}$;

\item[(a.3)] $\mathbf E^{l+1}  =  \mathbf E^{l} + \mathbf{M} \mathbf Y^{(k,l+1)} - \mathbf C^{l+1}$.
\end{itemize}
until $||\mathbf Y^{(k,l+1)}-\mathbf Y^{(k,l)}||\le \epsilon$. Note that in the above steps, $\mathbf C$ and $\mathbf E$ are two auxiliary matrices, $\mathbf C^0=\mathbf E^0=\mathbf 0$, $\mathbf Y^{(k,0)}=\mathbf Y^{(k)}$, $\mathbf F^{(k)}=\mathbf{P}^{(k)} - \mathbf{B}^{(k)}$, and $\epsilon$ is a predefined error tolerance. The problem in (a.1) is a least-squares problem and can be easily solved through the following sparse linear system,
\begin{equation}\label{eq:LSys}
\mathbf L \mathbf Y = r \mathbf D^{1/2} \mathbf F^{(k)}+\lambda \mathbf M^{\text{T}}(\mathbf C^{l} - \mathbf E^l),
\end{equation}
where $\mathbf L = \lambda \mathbf S +r\mathbf D$, and $\mathbf S (= \mathbf M^\text{T} \mathbf M$) can be explicitly defined as
\[
\mathbf{S} =
\begin{pmatrix}
  \underset{j\neq 1} \sum W_{1j}^2 & -W_{12}^2 & \cdots & -W_{1n}^2\\
 -W_{21}^2 &  \underset{j\neq 2} \sum W_{2j}^2 & \cdots & -W_{2n}^2\\
 \vdots & \vdots & \ddots & \vdots\\
 -W_{n1}^2 & -W_{n2}^2 & \cdots & \underset{j\neq n} \sum W_{nj}^2
\end{pmatrix},
\]
where $W_{ij}=0$ if there exists no connection between $x_i$ and $x_j$.

Matrix $\mathbf L$ has the following two important characteristics: 1) $\mathbf L$ stays the same across iterations and only the right hand side of (\ref{eq:LSys}) evolves; 2) $\mathbf L$ is sparse, symmetric and positive definite as it is a Hermitian diagonally dominant matrix with real positive diagonal entries. So $\mathbf L$ only needs to be decomposed once for all iterations using a sparse matrix solver. In practice, we use Cholesky matrix decomposition to factorize matrix $\mathbf L$: $\mathbf L = \mathbf{\Lambda} \mathbf{\Lambda}^\text{T}$ where $\mathbf{\Lambda}$ is a lower triangular matrix. Thus (\ref{eq:LSys}) is equivalent to two triangular systems,
\begin{eqnarray}\label{eq:TSys}
\nonumber \mathbf \Lambda \mathbf{\widetilde{Y}} & = & r \mathbf D^{1/2} \mathbf F^{(k)}+\lambda \mathbf M^{\text{T}}(\mathbf C^{l} - \mathbf E^l), \\
\mathbf \Lambda^\text{T} \mathbf Y & = & \mathbf{\widetilde{Y}}.
\end{eqnarray}

During each iteration, the extremely efficient forward and backward substitution method can be used to solve the sparse linear system in (\ref{eq:LSys}). This is the main reason for our choice of the Splitting Bregman algorithm in solving our $L_{1,1}$-regularized optimization.

In the conference version, we use the mldivide function in MATLAB, which performs Cholesky decomposition internally without returning the triangular matrix. Thus, we had to unnecessarily perform Cholesky decomposition in every iteration when solving (\ref{eq:LSys}). In this work, we perform Cholesky decomposition to factorize matrix $\mathbf L$ only once and save the resulting triangular matrix. This allows us to solve two triangular systems only in every iteration, as shown in (\ref{eq:TSys}). This gives rise to a much faster algorithm. 
In a test run on 200 sample images from the BSDS500 dataset, the average computation time of swPFE$_1$ drops from 233 seconds to 60 seconds.

The problem in (a.2) has a closed form solution called shrinkage operation defined as follows,
\begin{equation}
  \mathbf C^{l+1} =  \mbox{Shrink}(\mathbf{M} \mathbf Y^{(k,l+1)} + \mathbf E^{l}, 1/\lambda).
\end{equation}
Suppose $\mathbf{Z} =\mbox{Shrink}(\mathbf{X},\gamma)$. Then $Z_{ij} = \mbox{sign}(X_{ij}) \max(|X_{ij}|-\gamma, 0)$.


\section{$L_{1,p}$-regularized minimization}\label{appendII}
Suppose $\mathbf M = [\mathbf m_1,\cdots,\mathbf m_t]^{\text{T}}$ where $\mathbf m_i^{\text{T}}$ is the $i$-th row of $\mathbf M$. The problem in (\ref{eq:PFEoptL1p}) can be rewritten as,
\begin{equation} \label{eq:PFEoptL1p_new}
 \mathbf Y  =  \arg \underset{\mathbf Y} \min \sum_{i=1}^t \|\mathbf m_i^{\text{T}} \mathbf Y\|_{1}^{p}+ \Psi(\mathbf Y)
\end{equation}

According to \cite{ochs2015iteratively}, the majorization-minimization algorithm can be applied to solve problem (\ref{eq:PFEoptL1p}) in the form of the following iterative algorithm which makes the energy value monotonically decrease when $0<p<1$. 
\begin{itemize}
\item[\circled{1}] Solve the following nonsmooth $L_{1,1}$-regularized problem,
\begin{eqnarray} \label{eq:OptL11}
\nonumber \mathbf Y^{(k+1)} & = & \arg \underset{\mathbf Y} \min \sum_{i=1}^t w_{i}^{k}\|\mathbf m_i^{\text{T}} \mathbf Y\|_{1}+ \Psi(\mathbf Y) \\
& = & \arg \underset{\mathbf Y} \min \|\mathbf{\hat{M}}^{(k)} \mathbf Y\|_{1,1}+ \Psi(\mathbf Y),
\end{eqnarray}
where $\mathbf{\hat{M}}^{(k)} = [w_{1}^{k} \mathbf m_1,\cdots, w_{t}^{k} \mathbf m_t]^{\text{T}} $ and $w_{i}^{0}=1$.
\item[\circled{2}] Update the coefficient $w_{i}^{k}$:
\begin{equation}\label{eq:reweighting}
\nonumber w_{i}^{k+1} =p \; \mbox{max}(\|\mathbf m_i^{\text{T}} \mathbf Y ^{(k+1)}\|_{1},\varepsilon)^{p-1},
\end{equation}
where $\varepsilon$ is a constant used for avoiding an infinite value.
\end{itemize}
Again the $L_{1,1}$-regularized problem in (\ref{eq:OptL11}) can be solved using the splitting Bregman iterations described in appendix~\ref{appendI}.

\end{document}